\documentclass[10pt,twocolumn,letterpaper]{article}

\usepackage{cvpr}
\usepackage{times}
\usepackage{epsfig}
\usepackage{graphicx}
\usepackage{amsmath}
\usepackage{amssymb}
\usepackage{bm}
\usepackage{paralist}
\usepackage{subcaption}
\usepackage{cite}

\def\VEC#1{\bm{#1}} 

\usepackage[pagebackref=true,breaklinks=true,letterpaper=true,colorlinks,bookmarks=false]{hyperref}

\cvprfinalcopy % *** Uncomment this line for the final submission

\def\cvprPaperID{179} % *** Enter the CVPR Paper ID here

\ifcvprfinal\fi
\begin{document}

%%%%%%%%% TITLE
%\title{A Linear One-point Algorithm for Extrinsic Calibration of Kaleidoscopic Imaging System}
\title{A Linear Extrinsic Calibration of Kaleidoscopic Imaging System \\ from Single 3D Point}

% TODO
% - takahashi16mirror を参考文献に追加

\author{Kosuke Takahashi \quad Akihiro Miyata\thanks{Present affiliation: Nara Institute of Science and Technology.} \quad Shohei Nobuhara \quad Takashi Matsuyama\\
Kyoto University\\ 
%Yoshidahonmachi, Sakyo, Kyoto, Japan\\
{\tt\small \{takahasi,miyata,nob,tm\}@vision.kuee.kyoto-u.ac.jp}
%\and
}

\maketitle
%\thispagestyle{empty}

%%%%%%%%% ABSTRACT
\begin{abstract}
This paper proposes a new extrinsic calibration of kaleidoscopic imaging system by estimating normals and distances of the mirrors. The problem to be solved in this paper is a simultaneous estimation of all mirror parameters consistent throughout multiple reflections. Unlike conventional methods utilizing a pair of direct and mirrored images of a reference 3D object to estimate the parameters on a per-mirror basis, our method renders the simultaneous estimation problem into solving a linear set of equations.
The key contribution of this paper is to introduce a linear estimation of multiple mirror parameters from kaleidoscopic 2D projections of a single 3D point of unknown geometry. Evaluations with synthesized and real images demonstrate the performance of the proposed algorithm in comparison with conventional methods.
\end{abstract}

%%%%%%%%% BODY TEXT

\section{Introduction}
%\footnote[0]{A reference implementation is available at \url{http://vision.kuee.kyoto-u.ac.jp/~nob/proj/kaleidoscope/}.}
Virtual multiple-view system with planar mirrors is a practical approach to realize a multi-view capture of a target by synchronized cameras with an identical intrinsic parameter, and it has been widely used for 3D shape reconstruction by stereo\cite{nane98stereo,goshtasby93design,gluckman2001catadioptric}, shape-from-silhouette\cite{huang06contour,forbes06shape,reshetouski11three}, structure-from-motion\cite{Ramalingam2011light},  structured-lighting\cite{lanman09surround,tahara15interference}, ToF\cite{nobuhara16single}, and also for reflectance analysis\cite{ihrke2012kaleidoscopic,tagawa12eight,inoshita13full}, for light-field imaging\cite{levoy04synthetic,sen05dual,fuchs12design}, \etc. 

\begin{figure}[t]
\centering
\includegraphics[width=0.9\linewidth]{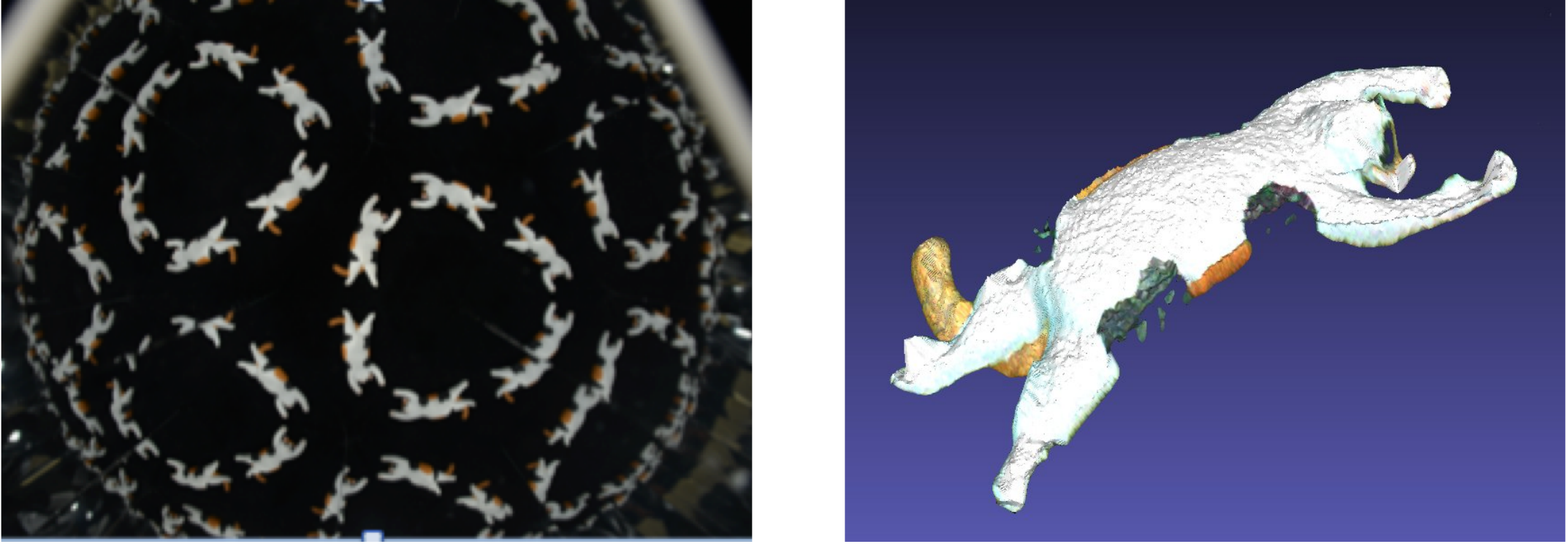}
\caption{Kaleidoscopic imaging system. Left: kaleidoscopic projection of a 3D \textit{cat} object. Right: a 3D reconstruction result.}
\label{fig:teaser}
\end{figure}

This paper is aimed at proposing a new extrinsic calibration of kaleidoscopic system with planar mirrors to provide an accurate and robust estimate of the mirror geometry for such applications (Figure \ref{fig:teaser}).

The problem addressed in this paper is to estimate all mirror parameters, \ie their normals and the distances from the camera, consistent throughout multiple reflections simultaneously in a linear manner. While conventional methods utilize a reference object of known geometry to estimate the mirror parameters on a per-mirror basis, the proposed method provides a linear solution of the mirror parameters from kaleidoscopic projections of a single 3D point without knowing its 3D geometry beforehand.

The key idea is to utilize the 2D projections of multiple reflections to form a linear system on the mirror parameters. While the 3D positions of multiple reflections of a 3D point is defined as a nonlinear function of the mirror parameters as described later in Eq \eqref{eq:hijhihj}, their 2D projections can be used as a linear constraint on the mirror parameters.

The rest of this paper is organized as follows. Section 2 reviews related studies on kaleidoscopic mirror calibrations. Section 3 defines the measurement model and Section 4 introduces a single mirror calibration algorithm from two pairs of projections based on the mirror-based binocular epipolar geometry\cite{ying13self}. Section 5 introduces our key contribution, a linear estimation of multiple mirror parameters from kaleidoscopic 2D projections of a single 3D point of unknown geometry. Section 6 evaluates the proposed method quantitatively and qualitatively in comparison with conventional methods, and Section 7 concludes the paper and outlines future work.
% The contribution of this paper is to introduce a linear estimation of multiple mirror parameters from kaleidoscopic 2D projections of a single 3D point of unknown geometry. Evaluations with synthesized and real images demonstrate the performance of the proposed algorithm in comparison with conventional methods.

% The contribution of this paper is twofold. Firstly, we introduce a two-point algorithm which provides a linear estimation of single mirror parameters from two pairs of real and mirrored points. Secondly, this algorithm is extended to estimate the mirror parameters from kaleidoscopic projections of a single 3D point.

\section{Related work}

In the context of kaleidoscopic imaging, Ihrke \etal\cite{ihrke2012kaleidoscopic} and Reshetouski and Ihrke\cite{reshetouski2013mirrors,reshetouski13discovering} have proposed a theory on modeling the chamber detection, segmentation, bounce tracing, shape-from-silhouette, \etc. In these studies, however, the geometric calibration of the mirrors is simply achieved by detecting chessboards first\cite{zhang1998}, and then by estimating the mirror normals and the distances from chessboard 3D positions in the camera frame.

By considering kaleidoscopic imaging as a system of observing reflections of a single object via different mirrors, another possible approach is to utilize calibration techniques from such mirrored observations\cite{sturm06how,kumar08simple,rodorigues10camera,hesch08mirror,takahashi12new,long15simplified}. While their original motivation is to estimate the 3D structure from its indirect views via mirrors, they can be used for calibrating the kaleidoscopic system by supposing the direct view were not available. For example, the orthogonality constraint on mirrored 3D points proposed by \cite{takahashi12new} can be considered as another approach for kaleidoscopic system calibration in \cite{ihrke2012kaleidoscopic,reshetouski2013mirrors}.

These conventional calibration approaches utilize 3D positions of a reference object and its reflections. That is, they first recover the 3D pose of the reference object from each of the virtual views, and then compute the mirror parameters from their 3D positions. While the first step and the second step can be done linearly, 3D pose estimation without nonlinear optimizations (\ie reprojection error minimization) is not robust to observation noise.

On the other hand, the proposed method directly estimates the mirror parameters linearly from kaleidoscopic projections of a single 3D point of unknown geometry, \ie without knowing its 3D position. Since our algorithm is based on a reprojection constraint, the result is as accurate as those with nonlinear optimizations.

\section{Kaleidoscopic imaging system}

\begin{figure}[t]
\centering
\includegraphics[width=0.8\linewidth]{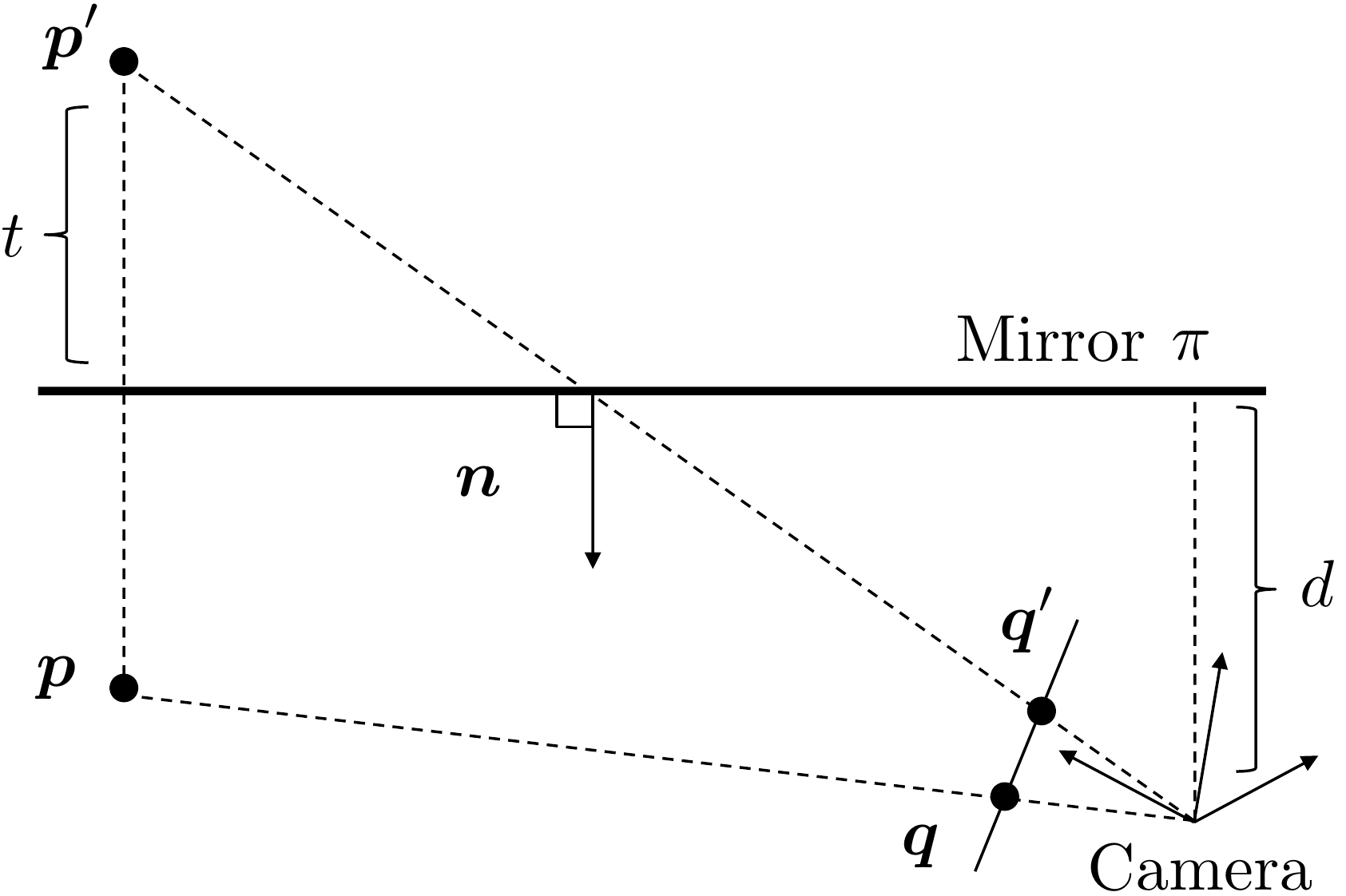}
\caption{Measurement model. A 3D point $\VEC{p}$ is reflected to $\VEC{p}'$ by a mirror $\pi$ of normal $\VEC{n}$ and distance $d$, and they are projected to $\VEC{q}$ and $\VEC{q}'$ respectively.}
\label{fig:mirror_model}
\end{figure}

\begin{figure}[t]
\centering
\includegraphics[width=0.6\linewidth]{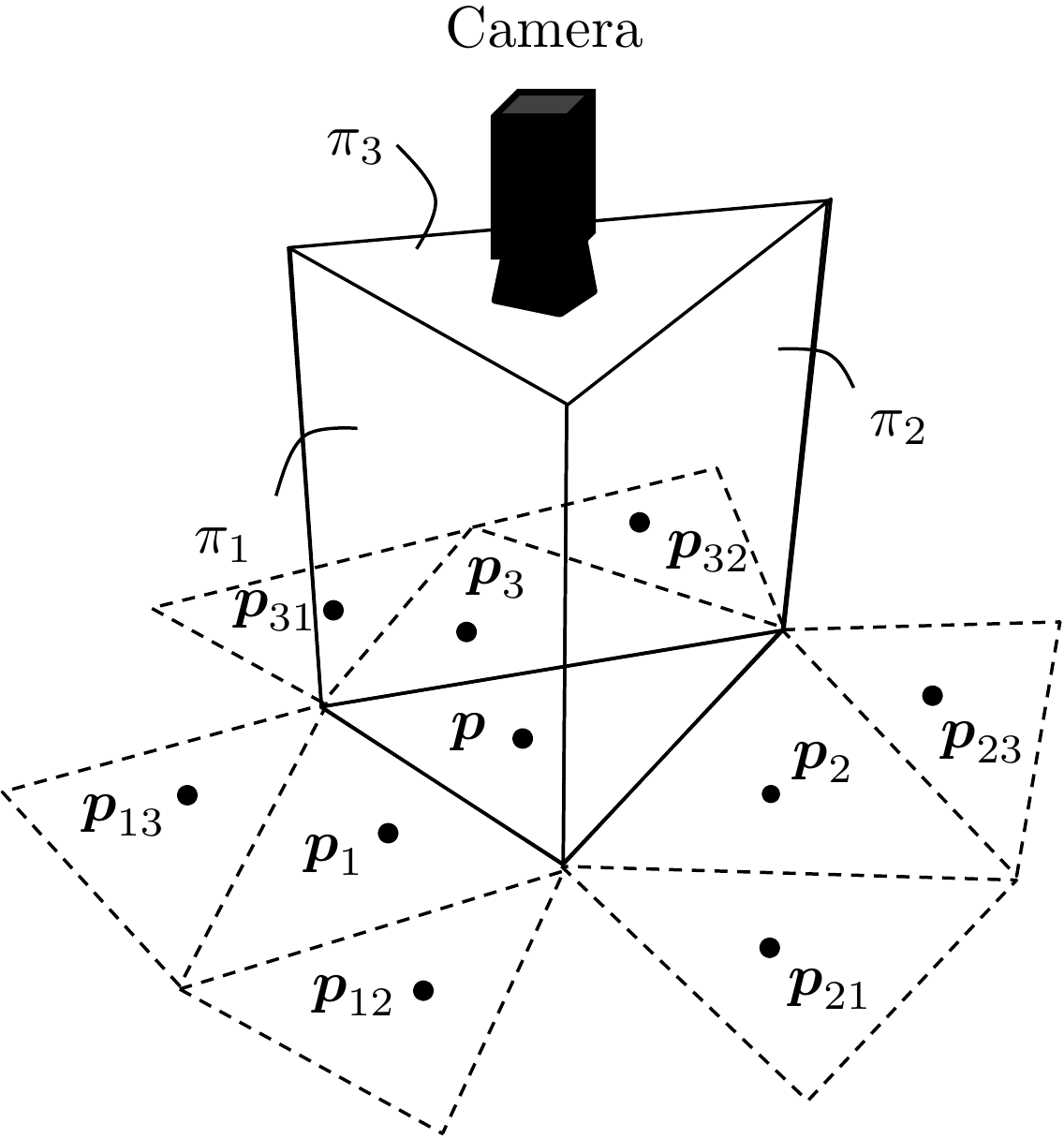}
\caption{Kaleidoscopic imaging system. A 3D point $\VEC{p}$ is reflected to $\VEC{p}_1$, $\VEC{p}_2$ and $\VEC{p}_3$ by the mirrors $\pi_1$, $\pi_2$ and $\pi_3$ respectively.}
\label{fig:kaleidoscope}
\end{figure}

\begin{figure}[t]
\centering
\includegraphics[width=1\linewidth]{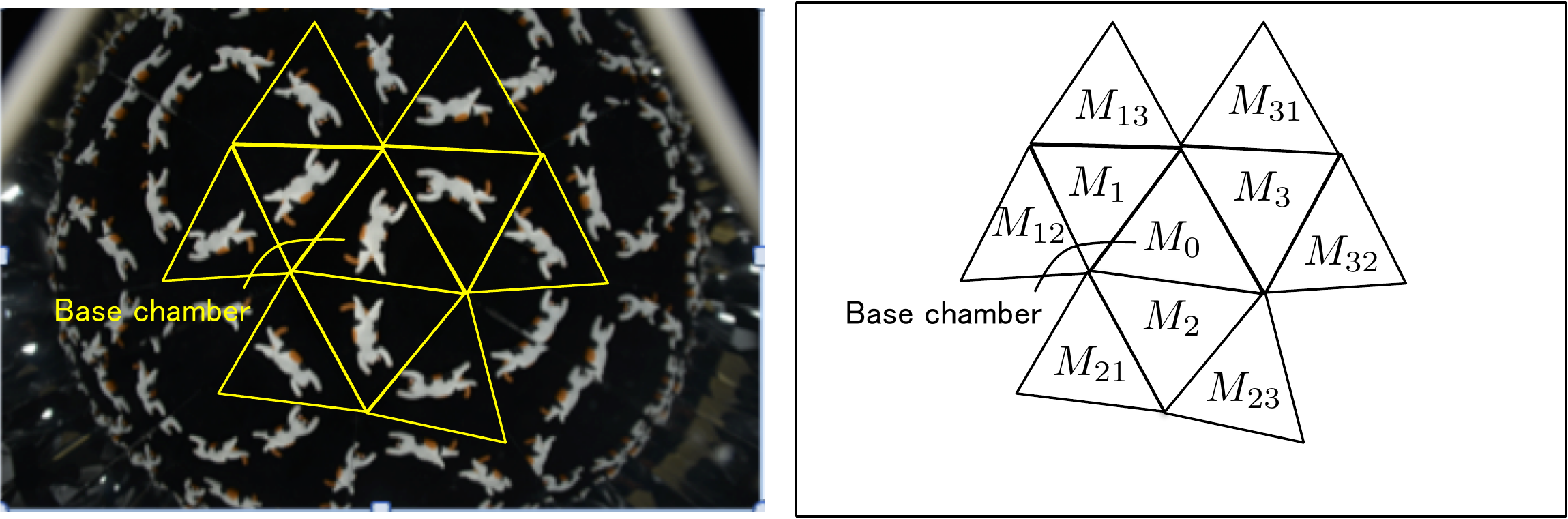}
\caption{Chamber arrangement}
\label{fig:chamber}
\end{figure}

Figure \ref{fig:mirror_model} illustrates the measurement model with a mirror. Let $\VEC{p}$ denote a 3D point in the camera coordinate system. The mirror $\pi$ of normal $\VEC{n}$ at distance $d$ from the camera generates its mirror as $\VEC{p}'$, and $\VEC{p}$ and $\VEC{p}'$ are captured as $\VEC{q}$ and $\VEC{q}'$ in the camera image
\begin{equation}
 \lambda \VEC{q} = A \VEC{p}, \quad \lambda' \VEC{q}' = A \VEC{p}', \label{eq:projection}
\end{equation}
where $A$ is the intrinsic matrix of the camera calibrated beforehand, and $\lambda$ and $\lambda'$ are the depths from the camera.

The 3D points $\VEC{p}$ and $\VEC{p}'$ satisfy
\begin{equation}
 \VEC{p} = \VEC{p}' + 2 t \VEC{n},
\end{equation}
where $t$ denotes the distance from $\VEC{p}$ to the mirror plane.
Also the projection of $\VEC{p}'$ to $\VEC{n}$ gives
\begin{equation}
 t + d = - \VEC{n}^\top \VEC{p}'.
\end{equation}
These two equations yield
\begin{equation}
 \VEC{p}=-2(\VEC{n}^\top \VEC{p}'+d)\VEC{n}+\VEC{p}',
\end{equation}
and can be rewritten as
\begin{equation}
 \tilde{\VEC{p}} = S \tilde{\VEC{p}}'
  = \begin{bmatrix}
     H  & -2d\VEC{n}\\
     \VEC{0}_{1\times 3} & 1
    \end{bmatrix}
    \tilde{\VEC{p}}', \label{eq:householder}
\end{equation}
where $H = I_{3{\times}3} - 2 \VEC{n} \VEC{n}^\top$ is a $3{\times}3$ Householder matrix, $\tilde{x}$ denotes the homogeneous coordinate of $x$, and $\VEC{0}_{m{\times}n}$ denotes the $m{\times}n$ zero matrix.

Kaleidoscopic imaging system utilize multiple mirrors to generate multiple viewpoints virtually (Figure \ref{fig:kaleidoscope}), and the images captured by the camera consist of \textit{chambers} corresponding to images captured by the real and the virtual cameras as shown in Figure \ref{fig:chamber}. Here we assume three mirrors system while our calibration can be adopted to other configurations.

Let $M_0$ denote the \textit{base} chamber corresponding to the direct view of the target. The three mirrors $\pi_1$, $\pi_2$ and $\pi_3$ generate first reflection chambers $M_1$, $M_2$ and $M_3$ respectively. These three mirrors also generate virtual mirrors $\pi_{ij}$ by mirroring $\pi_j$ by $\pi_i \: (i,j = 1,2,3, \: i \neq j)$. The matrices $S_{ij}$ and $H_{ij}$ of $\pi_{ij}$ are given by
\begin{equation}
\begin{split}
 S_{ij} & = S_i S_j, \\
 H_{ij} & = H_i H_j, \label{eq:hijhihj}
\end{split}
\end{equation}
and the camera observes the second reflection chamber $M_{ij}$ as the mirror of $M_j$ by $\pi_i$. The third and further reflections are defined by
\begin{equation}
 \Pi_{k=1}^m S_{i_k} \: (i_k = 1,2,3, \: i_{k} \neq i_{k+1}),
\end{equation}
where $m$ is the number of reflections.

The goal of our extrinsic calibration is to estimate the parameters $\VEC{n}_i$ and $d_i$ of the real mirror $\pi_i$ from projections of a single 3D point in the base chamber $M_0$ and its mirrors in $M_{i}$, $M_{ij}$, and so on.

%Notice that this paper employs a single camera and hence its local coordinate system serves as the world coordinate system.

\section{Single mirror calibration from projections of two 3D points}\label{sec:2pt}

Suppose the camera observes a 3D point of unknown geometry $\VEC{p}$. The mirror $\pi$ of matrix $S$ defined by the normal $\VEC{n}$ and the distance $d$ reflects $\VEC{p}$ to $\VEC{p}' = S\VEC{p}$ (Eq \eqref{eq:householder}).

Based on the epipolar geometry\cite{hartley00multiple,ying13self}, $\VEC{n}$, $\VEC{p}$ and $\VEC{p}'$ are coplanar and satisfy
\begin{equation}
 \left(\VEC{n} \times \VEC{p}\right)^\top \VEC{p}' = 0.
\end{equation}
By substituting $\VEC{p}$ and $\VEC{p}'$ by $\lambda A^{-1} \VEC{q}$ and $\lambda' A^{-1} \VEC{q}'$ respectively (Eq \eqref{eq:projection}), we obtain
\begin{equation}
 \VEC{q}^\top A^{-\top} [\VEC{n}]_{\times}^{\top} A^{-1} \VEC{q}' = 0, \label{eq:coplanarity}
\end{equation}
where $[\VEC{n}]_\times$ denotes the $3 \times 3$ skew-symmetric matrix representing the cross product by $\VEC{n}$ and this is the essential matrix of this mirror-based binocular geometry\cite{ying13self}.

By representing the normalized image coordinates of $\VEC{q}$ and $\VEC{q}'$ by $(x, y, 1)^\top = A^{-1}\VEC{q}$ and $(x', y', 1)^\top = A^{-1}\VEC{q}'$ respectively, Eq \eqref{eq:coplanarity} can be rewritten as
\begin{equation}
 \begin{pmatrix}
% x y' - x' y & x - x' & y - y'
 y - y' & x' - x & x y' - x' y
 \end{pmatrix}
 \VEC{n}
 = 0. \label{eq:coplanarity2}
\end{equation}
This equation allows estimating $\VEC{n}$ up to scale by using projections of more than or equal to two 3D points and their mirrors. Since $\VEC{n}$ is a unit vector, we can obtain a unique solution by assuming the mirror is front-facing to the camera.

It should be noted the distance $d$ from the camera to the mirror cannot be estimated since it is identical to the scale factor.

\section{Multiple mirrors calibration from kaleidoscopic projections of single 3D point}

\begin{figure}[t]
\centering
\includegraphics[width=0.6\linewidth]{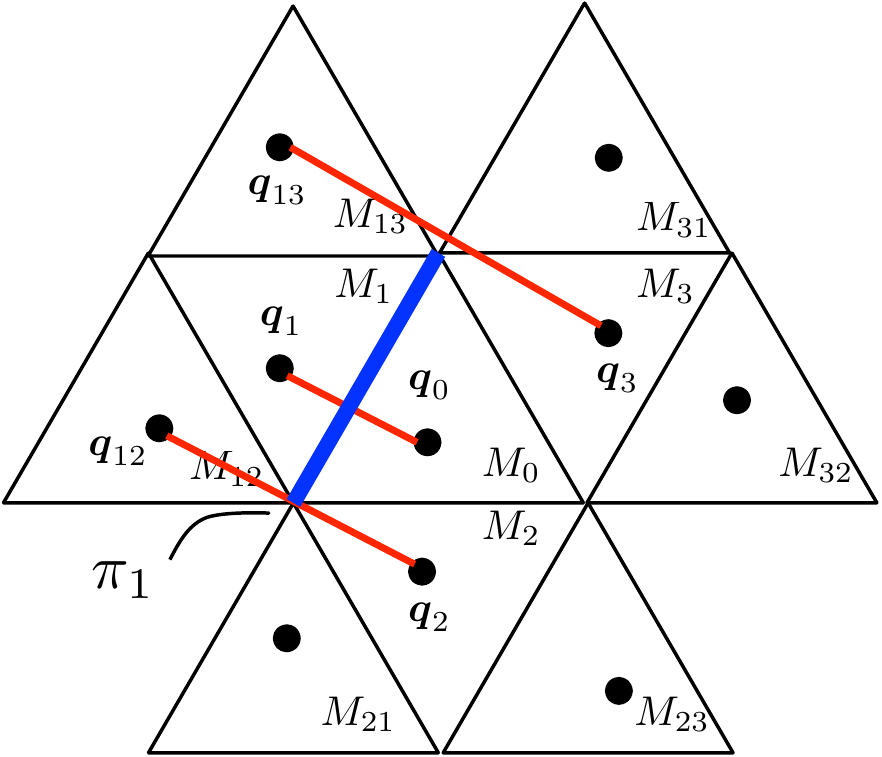}
\caption{Corresponding points. Three pairs $\langle\VEC{q}_0,\VEC{q}_1\rangle$, $\langle\VEC{q}_2,\VEC{q}_{12}\rangle$ and $\langle\VEC{q}_3,\VEC{q}_{13}\rangle$ (red) are available or mirror $\pi_1$ (blue)}
\label{fig:corr_proposed}
\end{figure}

This section introduces our linear algorithm which estimates the mirror normals and the distances from the kaleidoscopic projections of a single 3D point. Notice that the algorithm is first introduced by utilizing up to the second reflections, but they can be extended to third or further reflections intuitively as described later.

\subsection{Mirror normals $\VEC{n}_1$, $\VEC{n}_2$, and $\VEC{n}_3$}\label{sec:normal}

The algorithm in Section \ref{sec:2pt} realizes a mirror calibration on a per-mirror basis. That is, it can estimate the parameters of $\pi_1$, $\pi_2$ and $\pi_3$ independently. Furthermore, it can also estimate those of virtual mirrors such as $\pi_{13}$, $\pi_{23}$, and so forth.

However, such real mirror and virtual mirror parameters are not guaranteed to be consistent with each other and Eq \eqref{eq:hijhihj} does not hold strictly. This results in inconsistent triangulations in 3D geometry estimation for example.

Instead of such mirror-wise estimations, this section proposes a new linear algorithm which calibrates the kaleidoscopic mirror parameters simultaneously by observing a single 3D point in the scene.

Suppose a 3D point $\VEC{p}_0$ is projected to $\VEC{q}_0$ in the base chamber, and its mirror $\VEC{p}_i$ by $\pi_i$ is projected to $\VEC{q}_i$ in the chamber $M_i$. Likewise, the second mirror $\VEC{p}_{ij}$ by $\pi_{ij}$ is projected to $\VEC{q}_{ij}$ in the chamber $M_{ij}$, and so forth.

Here $\VEC{p}_1 = S_1 \VEC{p}_0$ indicates that $\VEC{q}_0$ and $\VEC{q}_1$ satisfy Eq \eqref{eq:coplanarity2} and provide a constraint for estimating the mirror normal $\VEC{n}_1$ of $\pi_1$ as described in Section \ref{sec:2pt}.
In addition, if $\VEC{p}_2 = S_2 \VEC{p}_0$ holds as well, we obtain $S_1 \VEC{p}_2 = S_1 S_2 \VEC{p}_0 \Leftrightarrow \VEC{p}_{12} = S_1 \VEC{p}_{2}$. That is, the projection $\VEC{q}_2$ corresponding to the first reflection $\VEC{p}_2$ and the projection $\VEC{q}_{12}$ corresponding to the second reflection $\VEC{p}_{12}$ also satisfy Eq \eqref{eq:coplanarity2} on $\VEC{n}_1$. Similarly, if $\VEC{p}_3 = S_3 \VEC{p}_0$ holds, $\VEC{q}_3$ and $\VEC{p}_{12}$ provides a linear constraint on $\VEC{n}_1$ as well.
% first reflections $\VEC{q}_2$ and $\VEC{q}_3$ also satisfy the same constraint with second reflections $\VEC{q}_{12}$ and $\VEC{q}_{13}$ by $\pi_1$ respectively. 
From these three constraints, $\VEC{n}_1$ can be estimated by solving
\begin{equation}
    \begin{bmatrix}
    y_0 - y_1 & x_1 - x_0 & x_0 y_1 - x_1 y_0 \\
    y_2 - y_{12} & x_{12} - x_2 & x_2 y_{12} - x_{12} y_2 \\
    y_3 - y_{13} & x_{13} - x_3 & x_3 y_{13} - x_{13} y_3 \\
    \end{bmatrix}
    \VEC{n}_1 = \VEC{0}_{3{\times}1}. \label{eq:n1}
\end{equation}
Similarly, $\VEC{n}_2$ and $\VEC{n}_3$ can be estimated by solving
\begin{equation}
    \begin{bmatrix}
    y_0 - y_2 & x_2 - x_0 & x_0 y_2 - x_2 y_0 \\
    y_3 - y_{23} & x_{23} - x_3 & x_3 y_{23} - x_{23} y_3 \\
    y_2 - y_{21} & x_{21} - x_1 & x_1 y_{21} - x_{21} y_1 \\
    \end{bmatrix}
    \VEC{n}_2 = \VEC{0}_{3{\times}1}, \label{eq:n2}
\end{equation}
and
\begin{equation}
    \begin{bmatrix}
    y_0 - y_3 & x_3 - x_0 & x_0 y_3 - x_3 y_0 \\
    y_1 - y_{31} & x_{31} - x_1 & x_1 y_{31} - x_{31} y_1 \\
    y_2 - y_{32} & x_{32} - x_2 & x_2 y_{32} - x_{32} y_2
    \end{bmatrix}
    \VEC{n}_3 = \VEC{0}_{3{\times}1}. \label{eq:n3}
\end{equation}

An important observation in this simple algorithm is the fact that (1) this is a linear algorithm while it utilizes multiple reflections, and (2) the estimated normals $\VEC{n}_1$, $\VEC{n}_2$ and $\VEC{n}_3$ are enforced to be consistent with each other while they are computed on a per-mirror basis apparently.

The first point is realized by using not the multiple reflections of a 3D position but their 2D projections. Intuitively a reasonable formalization of kaleidoscopic projection is to define a real 3D point in the scene, and then to express each of the projections of its reflections by Eq \eqref{eq:householder} coincides with the observed 2D position as introduced in Section \ref{sec:ba} later. This expression, however, is nonlinear in the normals $\VEC{n}_i \: (i=1,2,3)$ (\eg $\VEC{p}_{12} = S_1 S_2 \VEC{p}_0$). On the other hand, projections of such multiple reflections can be associated as a result of single reflection by Eq \eqref{eq:coplanarity2} directly (\eg $\VEC{n}_1$ with $\VEC{q}_{12}$ and $\VEC{q}_2$ as the projections of $\VEC{p}_{12}$ and $S_2\VEC{p}_0$ respectively). As a result, we can utilize 2D projections of multiple reflections in the linear systems above.

This explains the second point as well. The above constraint on $\VEC{q}_{12}$, $\VEC{q}_2$ and $\VEC{n}_1$ in Eq \eqref{eq:n1} assumes $\VEC{p}_2 = S_2\VEC{p}_0$ being satisfied, and it is enforced by $(A^{-1}\VEC{q}_2 \times A^{-1}\VEC{q}_0)^\top \VEC{n}_2 = 0$ in the first row of Eq \eqref{eq:n2}. Inversely, on estimating $\VEC{n}_1$ by Eq \eqref{eq:n1}, it enforces $\VEC{p}_1 = S_1\VEC{p}_0$ for Eqs \eqref{eq:n2} and \eqref{eq:n3}.

It should be noted that this algorithm can be extended to third or further reflections intuitively. For example, if $\VEC{p}_{23}$ and its reflection by $\pi_1$ is observable as $\lambda_{123}\VEC{q}_{123} = A \VEC{p}_{123} = A S_1 \VEC{p}_{23}$, then it provides
\begin{equation}
    \left( y_{23} - y_{123}, x_{23} - x_{123}, x_{23} y_{123} - x_{123} y_{23} \right) \VEC{n}_1 = 0,
\end{equation}
and can be integrated with Eq \eqref{eq:n1}.

% The expression $\VEC{q}_{12}$, $\VEC{q}_2$ as the projection of $S_2\VEC{p}_0$, and $\VEC{n}_1$
% the estimated normals $\VEC{n}_1$, $\VEC{n}_2$ and $\VEC{n}_3$ are consistent for describing reflections by virtual mirrors while they are estimated on a mirror-wise manner apparently.

% For example $\VEC{p}_{12}$ is the reflection of $\VEC{p}_1$ by the mirror $\pi_{12}$ and therefore $\VEC{p}_1$, $\VEC{p}_{12}$ and $\VEC{n}_{12}$ should be coplanar:
% \begin{equation}
%  \begin{pmatrix}
%  x_1 y_{12} - x_{12} y_1 & x_1 - x_{12} & y_1 - y_{12}
%  \end{pmatrix}
%  \VEC{n}_{12}
%  = 0.
% \end{equation}
% While this coplanarity constraint does not explicitly appear in the above equations, this is enforced indirectly as follows.

% % (I - 2 n1 n1) n2, q1, q12　
% ...

% This coplanarity is invariant to the Householder transform and hence it holds after applying $S_1$. That is, 
% \begin{align}
%      & \left(\VEC{n}_{12} \times \VEC{p}_1 \right)^\top \VEC{p}_{12} = 0, \\
%     \Leftrightarrow & \left((S_1\VEC{n}_{12}) \times (S_1\VEC{p}_1) \right)^\top (S_1\VEC{p}_{12}) = 0, \\
%     \Leftrightarrow & \left(\VEC{n}_{2} \times \VEC{p}_0 \right)^\top \VEC{p}_{2} = 0,
% \end{align}

% they should satisfy
% \begin{equation}
%  \begin{pmatrix}
%  x_1 y_{12} - x_{12} y_1 & x_1 - x_{12} & y_1 - y_{12}
%  \end{pmatrix}
%  \VEC{n}_{12}
%  = 0.
% \end{equation}

\subsection{Mirror distances $d_1$, $d_2$, and $d_3$}

Once the mirror normals $\VEC{n}_1$, $\VEC{n}_2$, and $\VEC{n}_3$ are given linearly, the mirror distances $d_1$, $d_2$, and $d_3$ can also be estimated linearly as follows.

%Since the global scale of the scene cannot be determined from 2D observations, $d_1$ can be assumed to be $d_1 = 1$ without loss of generality.

\paragraph{Kaleidoscopic reprojection constraint}

The perspective projection Eq \eqref{eq:projection} indicates that a 3D point $\VEC{p}_i$ and its projection $\VEC{q}_i$ should satisfy the collinearity constraint:
\begin{equation}
    (A^{-1} \VEC{q}_i) \times \VEC{p}_i = \VEC{x}_i \times \VEC{p}_i = \VEC{0}_{3{\times}1}, \label{eq:cross_i}
\end{equation}
where $\VEC{x}_i = \begin{pmatrix} x_i & y_i & 1 \end{pmatrix}^\top$ is the normalized camera coordinate of $\VEC{q}_i$ as defined earlier.
Since the mirrored points $\VEC{p}_i \: (i=1,2,3)$ are then given by Eq \eqref{eq:householder} as
\begin{equation}
    \begin{split}
    \VEC{p}_i & %= S_i \VEC{p}_0 
    = H_i \VEC{p}_0 - 2 d_i \VEC{n}_i %\\
    %& = \lambda_0 H_i \VEC{x}_0 - 2 d_i \VEC{n}_i
    ,
    \label{eq:pi}
    \end{split}
\end{equation}
and we obtain
\begin{equation}
\begin{split}
    \VEC{x}_i \times \VEC{p}_i & =  \VEC{x}_i \times (H_i \VEC{p}_0 - 2 d_i \VEC{n}_i) \\
    & = [ \VEC{x}_i ]_\times 
    \begin{bmatrix}
    H_i
    &
    -2 \VEC{n}_i
    \end{bmatrix}
    \begin{bmatrix}
    \VEC{p}_0 \\
    d_i
    \end{bmatrix}\\
    & = \VEC{0}_{3{\times}1}.
\end{split}
\end{equation}

Similarly, the second reflection $\VEC{p}_{ij}$ is also collinear with its projection $\VEC{q}_{ij}$:
\begin{equation}
\begin{split}
    & (A^{-1}\VEC{q}_{ij}) \times \VEC{p}_{ij} \\
    %= & [ \VEC{x}_{ij} ]_\times (S_i \VEC{p}_{j}) \\
    = & [ \VEC{x}_{ij} ]_\times (H_i \VEC{p}_{j} - 2 d_i \VEC{n}_i) \\
    = & [ \VEC{x}_{ij} ]_\times \left(H_i \left( H_j \VEC{p}_0 - 2 d_j \VEC{n}_j \right) -2 d_i  \VEC{n}_i\right) \\
    = &
    [ \VEC{x}_{ij} ]_\times
    \begin{bmatrix}
    H_i H_j & 
    - 2 \VEC{n}_i &
    - 2 H_i \VEC{n}_j
    \end{bmatrix}
    \begin{bmatrix}
    \VEC{p}_0 \\
    d_i \\
    d_j
    \end{bmatrix} \\
    = & \VEC{0}_{3{\times}1}. \label{eq:cross_ij}
\end{split}
\end{equation}

By using these constraints, we obtain a linear system of $\VEC{p}_0$, $d_1$, $d_2$, and $d_3$:
\begin{equation}
\begin{split}
    &
    \begin{bmatrix}
        [\VEC{x}_0]_\times & \VEC{0}_{3{\times}1} & \VEC{0}_{3{\times}1} & \VEC{0}_{3{\times}1} \\
        h_{1} & -2 [ \VEC{x}_{1} ]_\times \VEC{n}_1 & \VEC{0}_{3{\times}1} & \VEC{0}_{3{\times}1} \\
        h_{2} & \VEC{0}_{3{\times}1} & -2 [ \VEC{x}_{2} ]_\times \VEC{n}_2 & \VEC{0}_{3{\times}1} \\
        h_{3} & \VEC{0}_{3{\times}1} & \VEC{0}_{3{\times}1} & -2 [ \VEC{x}_{3} ]_\times \VEC{n}_3 \\
        h'_{1,2} & -2 [ \VEC{x}_{12} ]_\times \VEC{n}_1 & -2 h''_{1,2} & \VEC{0}_{3{\times}1} \\
        h'_{2,1} & -2 h''_{2,1} & -2 [ \VEC{x}_{21} ]_\times \VEC{n}_2 & \VEC{0}_{3{\times}1} \\
        h'_{2,3} & \VEC{0}_{3{\times}1} & -2 [ \VEC{x}_{23} ]_\times \VEC{n}_2 & -2 h''_{2,3} \\
        h'_{3,2} & \VEC{0}_{3{\times}1} & -2 h''_{3,2} & -2 [ \VEC{x}_{32} ]_\times \VEC{n}_3 \\
        h'_{3,1} & -2 h''_{3,1} & \VEC{0}_{3{\times}1} & -2 [ \VEC{x}_{31} ]_\times \VEC{n}_3 \\
        h'_{1,3} & -2 [ \VEC{x}_{13} ]_\times \VEC{n}_1 & \VEC{0}_{3{\times}1} & -2 h''_{1,3} \\
    \end{bmatrix}
    \begin{bmatrix}
        \VEC{p}_0 \\
        d_1 \\
        d_2 \\
        d_3
    \end{bmatrix} \\
    = & 
    K \begin{bmatrix}
        \VEC{p}_0 \\
        d_1 \\
        d_2 \\
        d_3
    \end{bmatrix} = \VEC{0}_{30{\times}1}, \label{eq:reproj}
\end{split}
\end{equation}
where $h_{i} = [ \VEC{x}_{i} ]_\times H_i$, $h'_{i,j} = [ \VEC{x}_{ij} ]_\times H_i H_j$, $h''_{i,j} = [ \VEC{x}_{ij} ]_\times H_i \VEC{n}_j$.
By computing the eigenvector corresponding to the smallest eigenvalue of $K^\top K$, $(\VEC{p}_0, d_1, d_2, d_3)^\top$ can be determined up to a scale factor. In this paper, we choose the scale that normalizes $d_1 = 1$.

Notice that Eq \eqref{eq:reproj} apparently has 30 equations, but only 20 of them are linearly independent. This is simply because each of the cross products by Eqs \eqref{eq:cross_i} and \eqref{eq:cross_ij} has only two independent constraints by definition.

Also, as discussed in Section \ref{sec:normal}, the above algorithm can be extended to third or further reflections as well. For example, given the reflection of $\VEC{p}_{23}$ by $\pi_1$ as $\lambda_{123}\VEC{q}_{123} = A \VEC{p}_{123} = A S_1 \VEC{p}_{23}$, then it provides
\begin{equation}
    [\VEC{x}_{123}]_\times
    \begin{bmatrix}
    (H_1 H_2 H_3)^\top \\
    -2\VEC{n}_1^\top \\
    -2 (H_1 \VEC{n}_2)^\top \\
    -2 (H_1 H_2 \VEC{n}_3)^\top
    \end{bmatrix}^\top
    \begin{bmatrix}
        \VEC{p}_0 \\
        d_1 \\
        d_2 \\
        d_3
    \end{bmatrix}
    = \VEC{0}_{3{\times}1},
\end{equation}
and can be integrated with Eq \eqref{eq:reproj}.

\begin{figure}[t]
\centering
\includegraphics[width=0.9\linewidth]{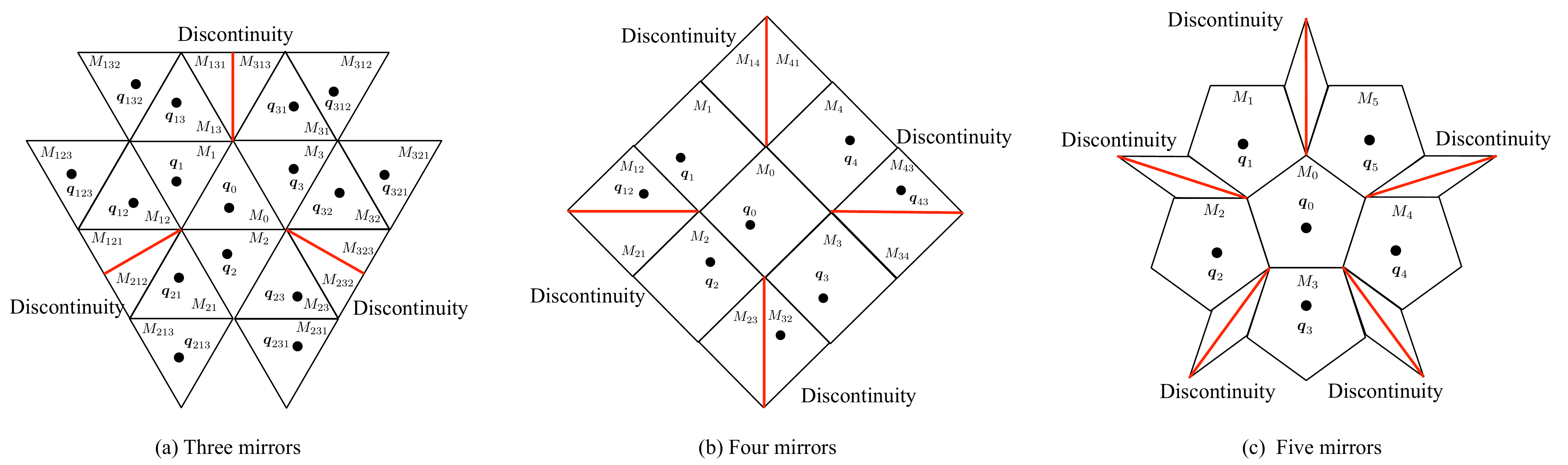}
\caption{Kaleidoscopic imaging system using (a) three, (b) four, and (c) five mirrors. Discontinuities (red lines) appear on the boundaries of overlapping chambers.}
\label{fig:various_configurations}
\end{figure}

Notice that our method works as long as the
second reflections by non-parallel mirrors are given regardless of the number of the mirrors. However, in cases of more than three mirrors, discontinuities are more likely to happen in general, and
finding the second reflections itself become difficult (Figure \ref{fig:various_configurations}).

% Notice that Eq \eqref{eq:reproj} minimizes reprojection errors, but this optimization is view-dependent. That is, the 3D point returned by this process tries to minimize reprojection errors but its 3D position has to lie on the viewing ray through $\VEC{x}_0$. In other words, this linear algorithm does not model position errors for $\VEC{x}_0$. The next introduces a bundle adjustment for our kaleidoscopic projection scenario.

\subsection{Kaleidoscopic bundle adjustment}\label{sec:ba}

%While the kaleidoscopic projection can be seen as a set of multi-view virtual cameras, optimizing their calibration parameters, \ie, rotations and translations, independently can break the reflection constraint. Instead, 

Once estimated the mirror normals $\VEC{n}_i$ and the distances $d_i \: (i=1,2,3)$ linearly, the triangulation from kaleidoscopic projections of a single 3D point can be given in a DLT manner by solving:
\begin{equation}
    K' \VEC{p}_0 = - K'' \VEC{d},
\end{equation}
as $\VEC{p}_0^\ast = - (K'^\top K')^{-1} K'^\top K'' \VEC{d}$, where $\VEC{d} = (d_1, d_2, d_3)^\top$, $K'$ is the $30{\times}3$ matrix corresponding to the first three columns of $K$:
\begin{equation}
\begin{split}
    & K' = \\ &
    \begin{bmatrix}
        [\VEC{x}_0]_\times^\top,
        h_{1}^\top,
        h_{2}^\top,
        h_{3}^\top,
        h^{'\top}_{1,2},
        h^{'\top}_{2,1},
        h^{'\top}_{2,3},
        h^{'\top}_{3,2},
        h^{'\top}_{3,1},
        h^{'\top}_{1,3}
    \end{bmatrix}^\top,
\end{split}
\end{equation}
and $K''$ is the $30{\times}3$ matrix corresponding to the 4th to 7th columns of $K$: 
\begin{equation}
    K'' = 
    \begin{bmatrix}
        \VEC{0}_{3{\times}1} & \VEC{0}_{3{\times}1} & \VEC{0}_{3{\times}1} \\
        -2 [ \VEC{x}_{1} ]_\times \VEC{n}_1 & \VEC{0}_{3{\times}1} & \VEC{0}_{3{\times}1} \\
        \VEC{0}_{3{\times}1} & -2 [ \VEC{x}_{2} ]_\times \VEC{n}_2 & \VEC{0}_{3{\times}1} \\
        \VEC{0}_{3{\times}1} & \VEC{0}_{3{\times}1} & -2 [ \VEC{x}_{3} ]_\times \VEC{n}_3 \\
        -2 [ \VEC{x}_{12} ]_\times \VEC{n}_1 & -2 h''_{1,2} & \VEC{0}_{3{\times}1} \\
        -2 h''_{2,1} & -2 [ \VEC{x}_{21} ]_\times \VEC{n}_2 & \VEC{0}_{3{\times}1} \\
        \VEC{0}_{3{\times}1} & -2 [ \VEC{x}_{23} ]_\times \VEC{n}_2 & -2 h''_{2,3} \\
        \VEC{0}_{3{\times}1} & -2 h''_{3,2} & -2 [ \VEC{x}_{32} ]_\times \VEC{n}_3 \\
        -2 h''_{3,1} & \VEC{0}_{3{\times}1} & -2 [ \VEC{x}_{31} ]_\times \VEC{n}_3 \\
        -2 [ \VEC{x}_{13} ]_\times \VEC{n}_1 & \VEC{0}_{3{\times}1} & -2 h''_{1,3} \\
    \end{bmatrix}.
\end{equation}

By reprojecting this $\VEC{p}_0^\ast$ to each of the chambers as
\begin{equation}
\begin{split}
    \lambda \hat{\VEC{q}}_0 & = A \VEC{p}_0^\ast, \\
    \lambda \hat{\VEC{q}}_{i} & = A S_i \VEC{p}_0^\ast \: (i=1,2,3), \\
    \lambda \hat{\VEC{q}}_{i,j} & = A S_i S_j \VEC{p}_0^\ast  \: (i,j=1,2,3, \: i \neq j),
\end{split}
\end{equation}
we obtain a reprojection error as
\begin{equation}
\begin{split}
    & \VEC{E}(\VEC{n}_1, \VEC{n}_2, \VEC{n}_3, d_1, d_2, d_3) \\
    & = 
    \begin{bmatrix}
        \VEC{q}_0 - \hat{\VEC{q}}_0,
        \VEC{e}_1,
        \VEC{e}_2,
        \VEC{e}_3,
        \VEC{e}'_{1,2},
        \VEC{e}'_{2,1},
        \VEC{e}'_{2,3},
        \VEC{e}'_{3,2},
        \VEC{e}'_{3,1},
        \VEC{e}'_{1,3}
    \end{bmatrix}^\top,
\end{split} \label{eq:reproj_error_vec}
\end{equation}
where $\VEC{e}_i = \VEC{q}_i - \hat{\VEC{q}}_i$ and $\VEC{e}'_{i,j} = \VEC{q}'_{i,j} - \hat{\VEC{q}}'_{i,j}$. By minimizing $|| \VEC{E}(\cdot) ||^2$ nonlinearly over $\VEC{n}_1, \VEC{n}_2, \VEC{n}_3, d_1, d_2, d_3$, we obtain a best estimate of the mirror normals and the distances.

\section{Evaluations}

\begin{figure}[t]
\centering
\includegraphics[width=0.6\linewidth]{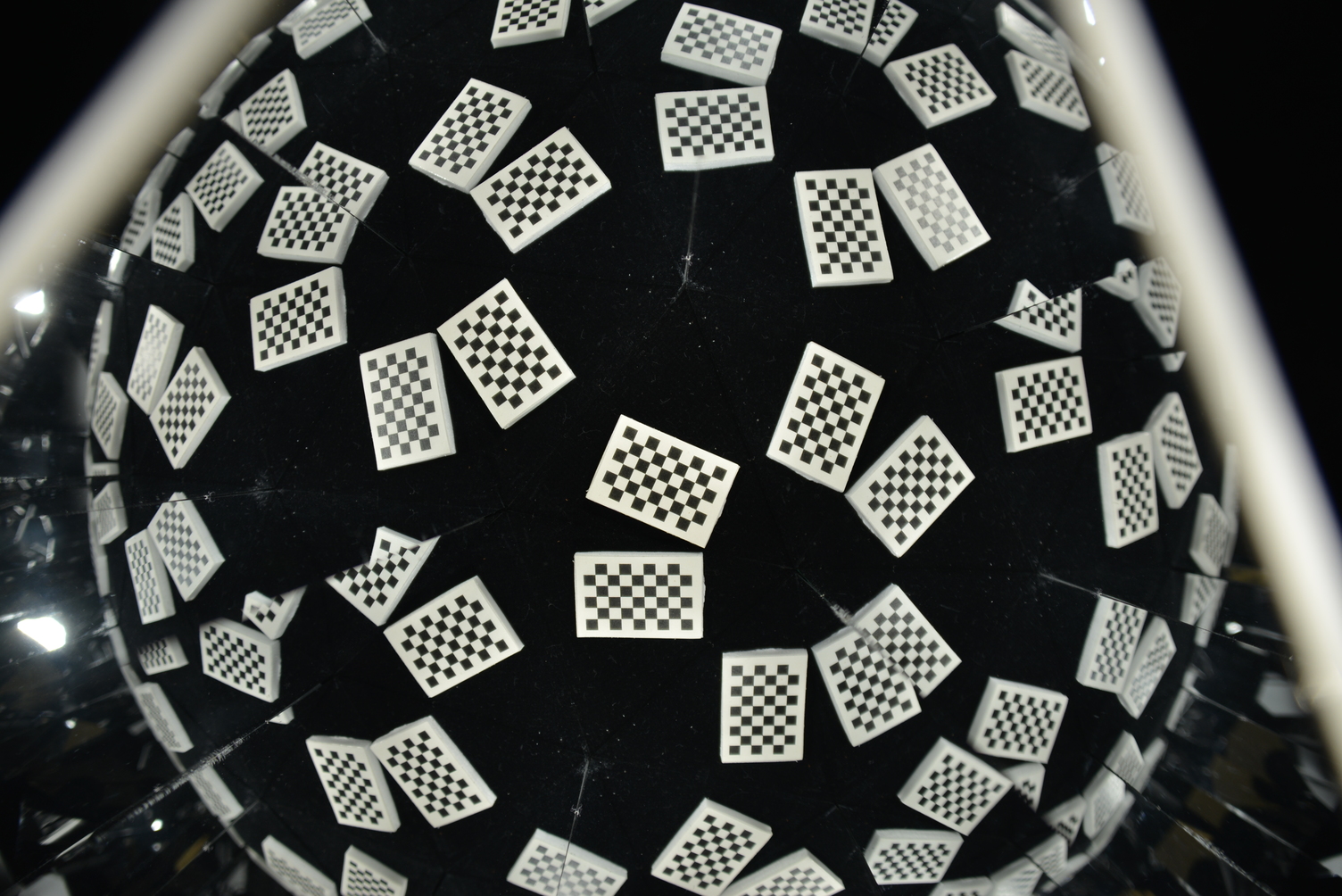}
\caption{A capture of a chessboard used as the reference object for conventional methods}
\label{fig:chessboard}
\end{figure}

\begin{figure}[t]
\centering
\includegraphics[width=0.6\linewidth]{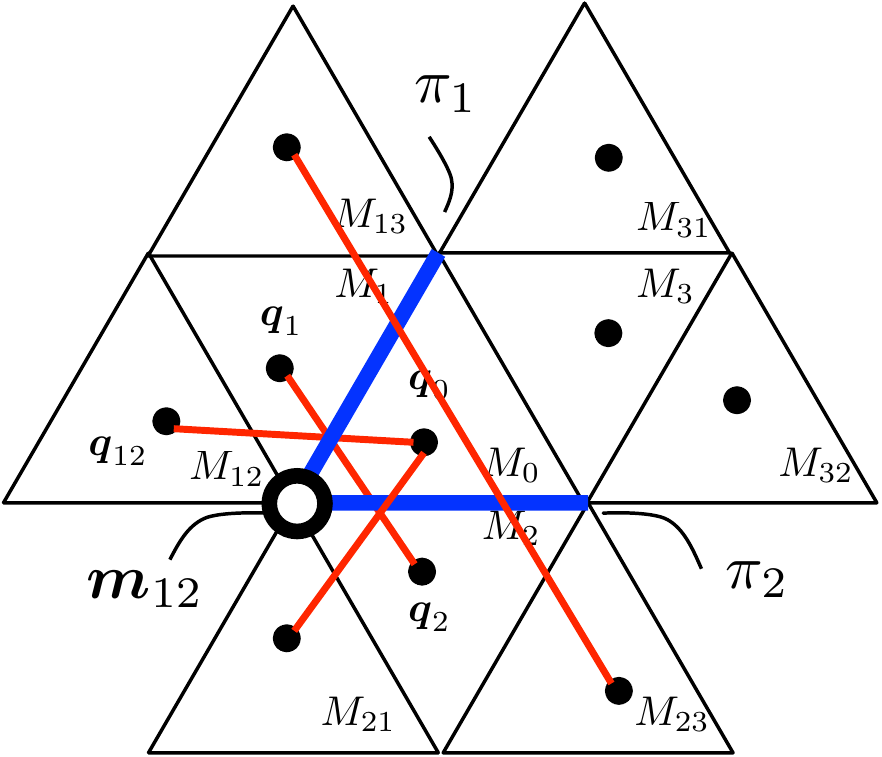}
\caption{Corresponding points for the orthogonality constraint\cite{takahashi12new}. Four pairs 
    $\langle\VEC{p}_{1}, \VEC{p}_{2}\rangle$, 
    $\langle\VEC{p}_{0}, \VEC{p}_{21}\rangle$,
    $\langle\VEC{p}_{12}, \VEC{p}_{0}\rangle$, and 
    $\langle\VEC{p}_{13}, \VEC{p}_{23}\rangle$ are available for the intersection $\VEC{m}_{12} = \VEC{n}_1 \times \VEC{n}_2$.}
\label{fig:corr_takahashi}
\end{figure}

To demonstrate the performance of the proposed algorithm, this section provides evaluations using synthesized and real images in comparison with the following two conventional algorithms both utilize a reference object of known geometry as shown in Figure \ref{fig:chessboard}.
\begin{description}
    \item[Baseline] 
    Since the 3D geometry of the reference object is known, the 3D positions of the real image $\VEC{p}_0^{(l)}$ and their reflections $\VEC{p}_i^{(l)}$ and $\VEC{p}_{i,j}^{(l)}$ can be estimated by solving PnP\cite{lepetit08epnp}. Here the superscript ${}^{(l)}$ indicates the $l$th landmark in the reference object. 
    Once $L$ such landmark 3D positions are given, then the mirror normals can be computed simply by
    \begin{equation}
    \begin{split}
        \VEC{n}_1 = \sum_l^L \VEC{l}_{1,2,3}^{(l)} / \left\|\sum_l^L \VEC{l}_{1,2,3}^{(l)}\right\|, \\
        \VEC{n}_2 = \sum_l^L \VEC{l}_{2,3,1}^{(l)} / \left\|\sum_l^L \VEC{l}_{2,3,1}^{(l)}\right\|, \\
        \VEC{n}_3 = \sum_l^L \VEC{l}_{3,1,2}^{(l)} / \left\|\sum_l^L \VEC{l}_{3,1,2}^{(l)}\right\|, \\
    \end{split}
    \end{equation}
    where $\VEC{l}_{i,j,k}^{(l)} = \VEC{p}_{i}^{(l)} - \VEC{p}_{0}^{(l)} + \VEC{p}_{ij}^{(l)} - \VEC{p}_{j}^{(l)} + \VEC{p}_{ik}^{(l)} - \VEC{p}_{k}^{(l)}$, and then the mirror distances can be computed by
    \begin{equation}
    \begin{split}
%        d_1 = \frac{1}{6L}\VEC{n}_1^\top\sum_l^L\left( \VEC{p}_{1}^{(l)} + \VEC{p}_{0}^{(l)} + \VEC{p}_{12}^{(l)} + \VEC{p}_{2}^{(l)} + \VEC{p}_{13}^{(l)} + \VEC{p}_{3}^{(l)}\right),\\
        d_1 = \frac{1}{6L}\VEC{n}_1^\top\sum_l^L\left( \sum_{i=0}^3 \left( \VEC{p}_{i}^{(l)} \right) + \VEC{p}_{12}^{(l)} + \VEC{p}_{13}^{(l)}\right),\\
        d_2 = \frac{1}{6L}\VEC{n}_2^\top\sum_l^L\left( \sum_{i=0}^3 \left( \VEC{p}_{i}^{(l)} \right) + \VEC{p}_{23}^{(l)} + \VEC{p}_{21}^{(l)}\right),\\
        %d_2 = \frac{1}{6L}\VEC{n}_2^\top\sum_l^L\left( \VEC{p}_{2}^{(l)} + \VEC{p}_{0}^{(l)} + \VEC{p}_{23}^{(l)} + \VEC{p}_{3}^{(l)} + \VEC{p}_{21}^{(l)} + \VEC{p}_{1}^{(l)}\right),\\
        d_3 = \frac{1}{6L}\VEC{n}_3^\top\sum_l^L\left( \sum_{i=0}^3 \left( \VEC{p}_{i}^{(l)} \right) + \VEC{p}_{31}^{(l)} + \VEC{p}_{32}^{(l)}\right).\\
        %d_3 = \frac{1}{6L}\VEC{n}_3^\top\sum_l^L\left( \VEC{p}_{3}^{(l)} + \VEC{p}_{0}^{(l)} + \VEC{p}_{31}^{(l)} + \VEC{p}_{1}^{(l)} + \VEC{p}_{32}^{(l)} + \VEC{p}_{2}^{(l)}\right).
    \end{split}
    \end{equation}
    Notice that the above PnP procedure requires a non-linear reprojection error minimization process in practice.
    
    \item[Takahashi \etal\cite{takahashi12new}]
    As pointed out by Takahashi \etal\cite{takahashi12new}, two 3D points $\VEC{p}_i$ and $\VEC{p}_j$ defined as reflections of a 3D point by different mirrors of normal $\VEC{n}_i$ and $\VEC{n}_j$ respectively satisfy an orthogonality constraint:
    \begin{equation}
        \left( \VEC{p}_i - \VEC{p}_j \right)^\top \left( \VEC{n}_i \times \VEC{n}_j \right) = \left( \VEC{p}_i - \VEC{p}_j \right)^\top \VEC{m}_{ij} = 0. \label{eq:orthogonality}
    \end{equation}
    As illustrated by Figure \ref{fig:corr_takahashi}, this constraint on $\VEC{m}_{12}$ holds for four pairs 
    $\langle\VEC{p}_{1}, \VEC{p}_{2}\rangle$, 
    $\langle\VEC{p}_{0}, \VEC{p}_{21}\rangle$,
    $\langle\VEC{p}_{12}, \VEC{p}_{0}\rangle$, and 
    $\langle\VEC{p}_{13}, \VEC{p}_{23}\rangle$ 
    as the reflections of $\VEC{p}_0$, $\VEC{p}_1$, $\VEC{p}_2$, and $\VEC{p}_3$ respectively. Similarly, 
    $\langle\VEC{p}_{2}, \VEC{p}_{3}\rangle$, 
    $\langle\VEC{p}_{21}, \VEC{p}_{31}\rangle$, 
    $\langle\VEC{p}_{0}, \VEC{p}_{32}\rangle$, and
    $\langle\VEC{p}_{23}, \VEC{p}_{0}\rangle$
    can be used for computing $\VEC{m}_{23} = \VEC{n}_2 \times \VEC{n}_3$, and
    $\langle\VEC{p}_{3}, \VEC{p}_{1}\rangle$,
    $\langle\VEC{p}_{31}, \VEC{p}_{0}\rangle$,
    $\langle\VEC{p}_{32}, \VEC{p}_{12}\rangle$, and
    $\langle\VEC{p}_{0}, \VEC{p}_{13}\rangle$
    can be used for $\VEC{m}_{31} = \VEC{n}_3 \times \VEC{n}_1$. Once obtained the intersection vectors $\VEC{m}_{12}$, $\VEC{m}_{23}$ and $\VEC{m}_{31}$, the mirror normals and the distances can be estimated linearly as described in \cite{takahashi12new}.
    
%     concatenating these three constraints for $L$ points, we obtain $3L$ linear equations for $\VEC{m}_{12} = \VEC{n}_1 \times \VEC{n}_2$. Once obtained $\VEC{m}_{23}$ and $\VEC{m}_{31}$ as well, the normals are given by 

% Substituting $\VEC{p}_i$ and $\VEC{p}_j$ by Eq \eqref{eq:pi} yields:
% \begin{equation}
% \begin{split}
%     & \left( \VEC{n}_i \times \VEC{n}_j \right)^\top \left( \VEC{p}_i - \VEC{p}_j \right) \\
%     = & \left( \VEC{n}_i \times \VEC{n}_j \right)^\top \left( \left(\lambda_0 \VEC{r}_i  - 2  d_i \VEC{n}_i\right) - \left(\lambda_0 \VEC{r}_j  - 2  d_i \VEC{n}_j\right) \right) \\
%     = & \left( \VEC{n}_i \times \VEC{n}_j \right)^\top \left( \lambda_0 \VEC{r}_i - \lambda_0 \VEC{r}_j   \right)
% \end{split}
% \end{equation}

% \begin{equation}
%     \left( \VEC{n}_i \times \VEC{n}_j \right)^\top 
%     \begin{bmatrix}
%     \VEC{r}_i - \VEC{r}_j & -2\VEC{n}_i & 2\VEC{n}_j
%     \end{bmatrix}
%     \begin{bmatrix}
%     \lambda_0 \\
%     d_i \\
%     d_j
%     \end{bmatrix}
%     = 0.
% \end{equation}

%     The constraint states that two reflections of a single landmark by different mirrors ($\VEC{p}_1$ and $\VEC{p}_2$ are reflections of $\VEC{p}_0$, for example) satisfy:
%     \begin{equation}
%         (VEC{p}_1 - \VEC{p}_2
%     \end{equation}
\end{description}

The following three error metrics are used in this section in order to evaluate the performance of the proposed method in comparison with the above-mentioned conventional approaches quantitatively. The average estimation error of normal $E_{\VEC{n}}$ measures the average angular difference from the ground truth by
\begin{equation}
    E_{\VEC{n}} = \frac{1}{3} \sum_{i=1}^3 \left|\cos^{-1}(\VEC{n}_i^\top \check{\VEC{n}}_{i}) \right|,
\end{equation}
where $\check{\VEC{n}}_i \: (i=1,2,3)$ denotes the ground truth of the normal $\VEC{n}_{i}$.
The average estimation error of distance $E_{d}$ is defined as the average $L_1$-norm to the ground truth:
\begin{equation}
    E_{d} = \frac{1}{3} \sum_{i=1}^{3} |d_i - \check{d}_{i}|,
\end{equation}
where $\check{d}_{i} \: (i=1,2,3)$ denotes the ground truth of the distance $d_{i}$.
Also, the average reprojection error $E_\mathrm{rep}$ is defined as:
\begin{equation}
    E_\mathrm{rep} = \frac{1}{10L} \sum_{l=1}^{L} \left| \VEC{E}^{(l)}(\VEC{n}_1, \VEC{n}_2, \VEC{n}_3, d_1, d_2, d_3) \right|,
\end{equation}
where $\VEC{E}^{(l)}(\cdot)$ denotes the reprojection error $\VEC{E}(\cdot)$ defined by Eq \eqref{eq:reproj_error_vec} at $l$th point.

\subsection{Quantitative evaluations with synthesized images}

% \begin{table}[t]
%     \centering
%     \includegraphics[width=\linewidth]{./fig/label.pdf}
%     \caption{Configurations}
%     \label{fig:label}
% \end{table}
\begin{figure*}[t]
    \centering
    \includegraphics[width=0.95\linewidth]{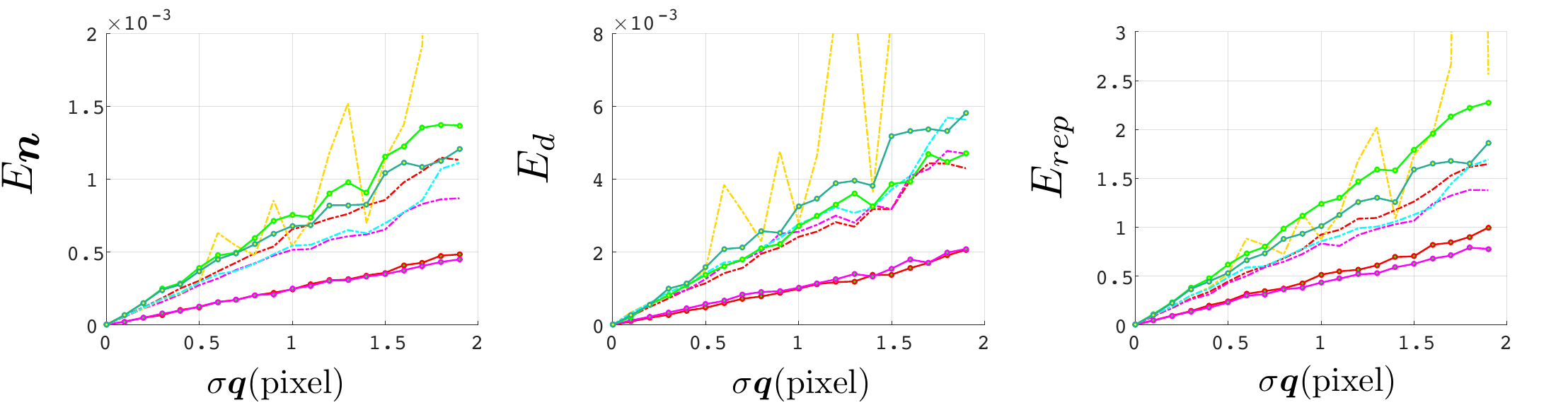}
    \caption{Estimation errors at different noise levels $\sigma_{\VEC{q}}$. Legends are provided in Table \ref{fig:label}.}
    \label{fig:simulation_noise}
%\end{figure*}
%\begin{figure*}[t]
%    \centering
    \includegraphics[width=0.95\linewidth]{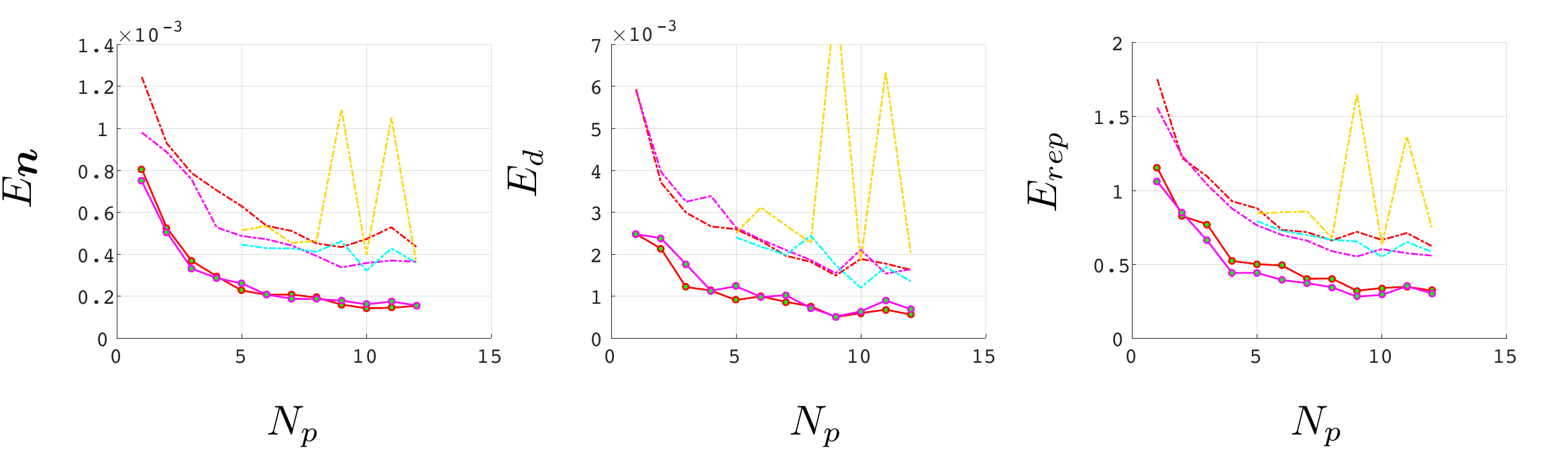}
    \caption{Estimation errors at different numbers of reference points $N_p$. Legends are provided in Table \ref{fig:label}.}
    \label{fig:simulation_point}
%\end{figure*}
%\begin{figure*}[t]
%    \centering
    \begin{minipage}[c]{0.62\linewidth}
    \includegraphics[width=\linewidth]{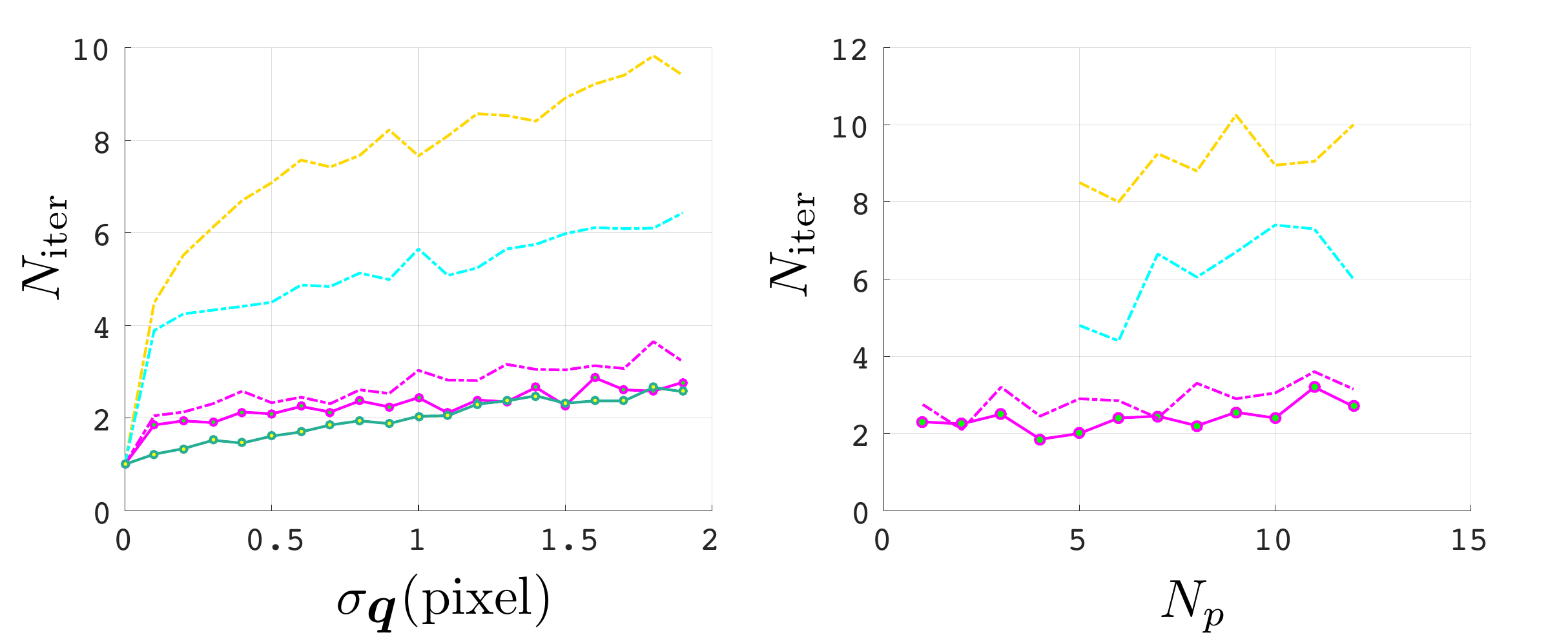}
    \caption{Number of iterations at different $\sigma_{\VEC{q}}$ with $N_p=5$ (left) and at different $N_p$ with $\sigma_{\VEC{q}}=1$ (right). Legends are provided in Table \ref{fig:label}.}
    \label{fig:simulation_iteration}
    \end{minipage}
    \begin{minipage}[c]{0.32\linewidth}
    \centering
    \includegraphics[width=\linewidth]{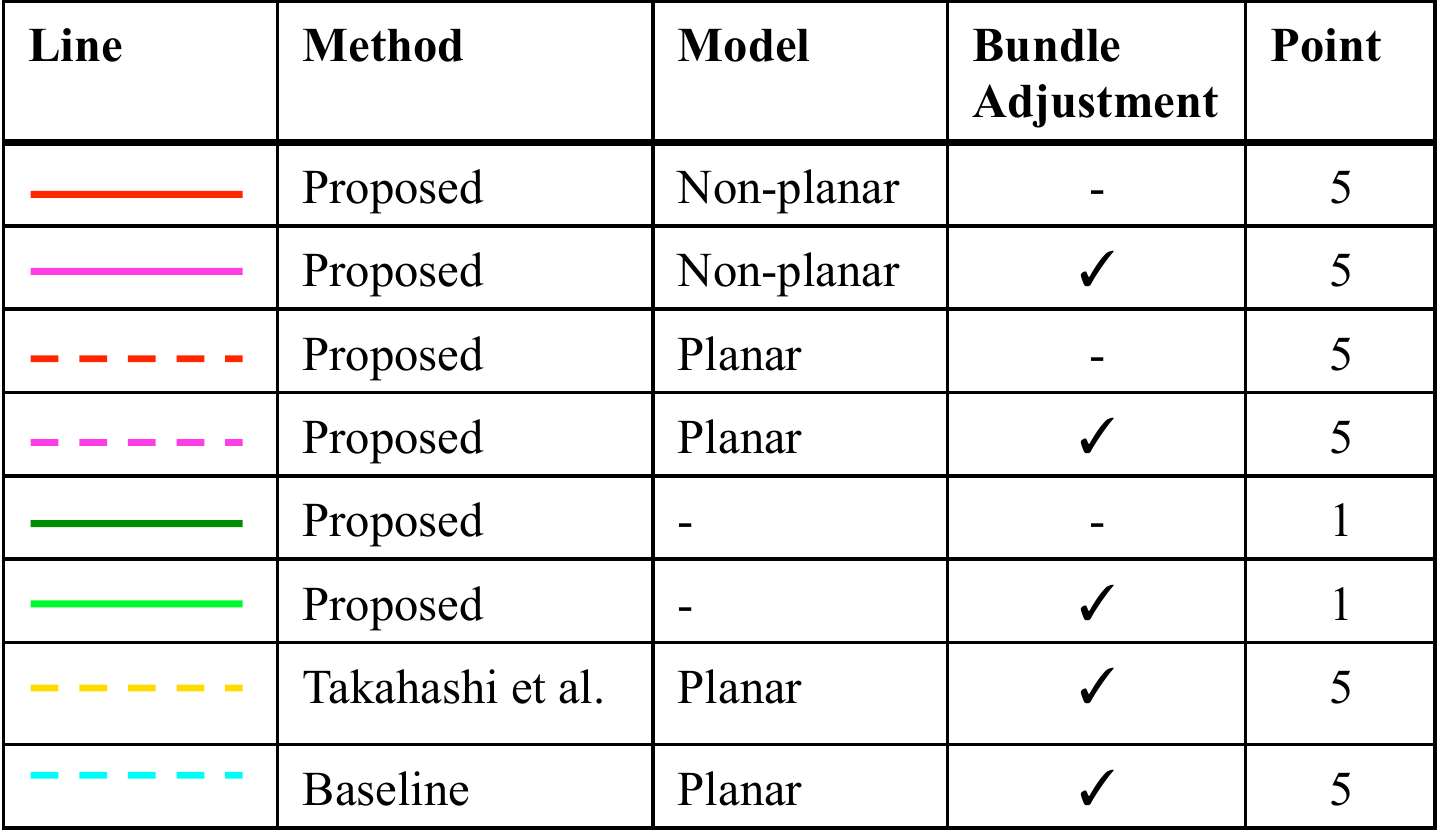}
    \caption{Configurations}
    \label{fig:label}
    \end{minipage}
\end{figure*}

This section provides a quantitative performance evaluation using synthesized dataset. A virtual camera and three mirrors are arranged according to the real setup (Figure \ref{fig:setup}). By virtually capturing 3D points simulating a reference object, the corresponding 2D kaleidoscopic projections used as the ground truth are generated first, and then random pixel noise is injected to them at each trial of calibration.

Figures \ref{fig:simulation_noise}, \ref{fig:simulation_point} and \ref{fig:simulation_iteration} report average estimation errors $E_{\VEC{n}}$, $E_d$, $E_\mathrm{rep}$ over 100 trials at different noise levels and different numbers of reference points. In these figures $\sigma_{\VEC{q}}$ denotes the standard deviation of zero-mean Gaussian pixel noise, $N_p$ denotes the number of 3D points used in the calibration, and $N_\mathrm{iter}$ denotes the number of iterations required by the kaleidoscopic bundle adjustment.
%In Figure \ref{fig:simulation_noise}, $N_p = 1$ for the proposed algorithm and $N_p = 5$ for the baseline method. In Figure \ref{fig:simulation_point}, $\sigma_{\VEC{q}} = 2.0$ for all methods.

As shown in Table \ref{fig:label}, the magenta and red lines denote the results by the proposed method with and without the non-linear optimization (Section \ref{sec:ba}). They use kaleidoscopic projections of non-planar random five 3D points, while the dashed red and magenta lines are the results with planar five points simulating the chessboard (Figure \ref{fig:chessboard}). The light and dark green lines are the results with a single 3D point generated randomly followed by the non-linear optimization or not.

The yellow and cyan dashed lines are the results by Takahashi \etal\cite{takahashi12new} and the baseline with the same five points for the red and magenta dashed lines. Notice that the baseline and Takahashi \etal\cite{takahashi12new} without the final non-linear optimization could not achieve comparable results (typically $E_\mathrm{rep} \gg 10$ pixel). Also these methods using 3D reference positions without applying non-linear refinement after a linear PnP\cite{lepetit08epnp} could not estimate valid initial parameters for the final non-linear optimization. Therefore, they are omitted in these figures. On the other hand, the final non-linear optimization for our method does not improve the result drastically. This is because our algorithm originally utilizes the reprojection error constraint.

From these results, we can conclude that (1) the proposed method can achieve comparable estimation linearly even with a single 3D point (dark green), and (2) the proposed method (red and magenta) with the same number 3D points used in the conventional methods (yellow and cyan) performs better, even without the final non-linear optimization.

Also in particular in the cases of $\sigma_{\VEC{q}} \ge 1$, we can observe Takahashi \etal (yellow) do not show robust behavior. This is because the method degenerates obviously if the intersection vectors $\VEC{m}_{12}$, $\VEC{m}_{23}$ and $\VEC{m}_{31}$ are parallel since the normal is recovered by $\VEC{n}_i = \VEC{m}_{ij} \times \VEC{m}_{ki}$. Therefore if the estimated 3D reference points by PnP return the intersection vectors close to such a singular configuration due to noise, then it will not perform robustly\cite{takahashi12new,agrawal13extrinsic}.

% if a vector $\VEC{p}_i^{(l)} - \VEC{p}_j^{(l)}$ and another vector $\VEC{p}_i^{(l')} - \VEC{p}_j^{(l')}$ used in Eq \eqref{eq:orthogonality} are parallel. This indicates that in the kaleidoscopic configuration as shown in Figure \ref{fig:chessboard}, chess corner pairs between two boards cannot provide sufficiently independent constraints.

% Also while it is not explicitly discussed in \cite{takahashi12new}, the method degenerates obviously if $\VEC{m}_{12}$, $\VEC{m}_{23}$ and $\VEC{m}_{31}$ are parallel since the normal is recovered by $\VEC{n}_i = \VEC{m}_{ij} \times \VEC{m}_{ki}$.

% From these two difficulties, \cite{takahashi12new} could not perform properly for the kaleidoscopic configuration.

\subsection{Qualitative evaluations with real images}

\begin{figure}[t]
  \centering
    \includegraphics[width=\linewidth]{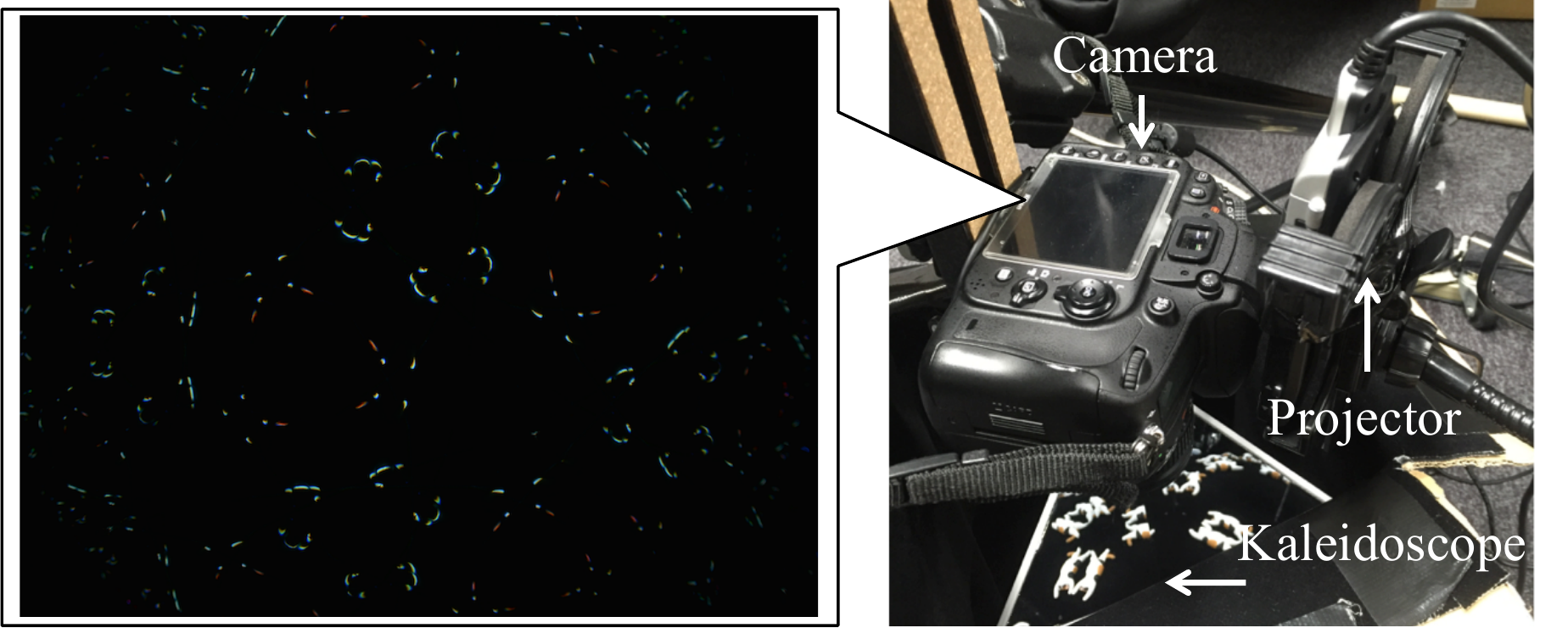}
    \caption{Capture setup}
    \label{fig:setup}
\end{figure}

\begin{figure}[t]
  \centering
    \includegraphics[width=0.6\linewidth]{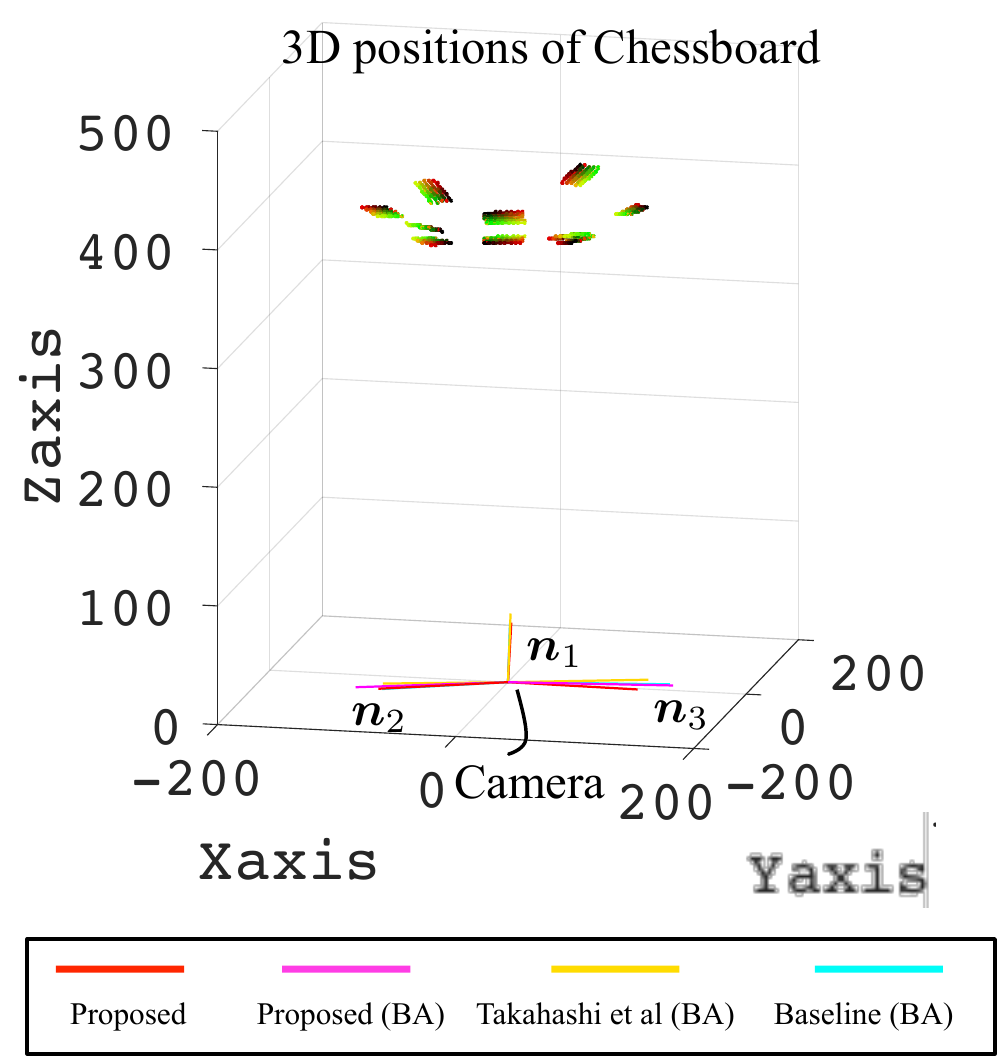}
    \caption{Calibration results. The colored lines in the bottom illustrate $d \VEC{n}$ (\ie the foot of perpendicular from the camera center) of each mirrors. The 10 patterns in the top illustrate the 3D points estimated by PnP.}
    \label{fig:real_chess}
\end{figure}

Figure \ref{fig:setup} shows our kaleidoscopic capture setup. The intrinsic parameter $A$ of the camera (Nikon D600, $6016{\times}4016$ resolution) is calibrated beforehand\cite{zhang1998}, and it observes the target object \textit{cat} (about $4\times5\times1$ cm) with three planar first surface mirrors. The projector (MicroVision SHOWWX+ Laser Pico Projector, $848{\times}480$ resolution) is used to cast line patterns to the object for simplifying the correspondence search problem in a light-sectioning fashion (Figure \ref{fig:setup} left), and the projector itself is not involved in the calibration \wrt the camera and the mirrors.

\begin{figure}[t]
  \centering
    \includegraphics[width=0.6\linewidth]{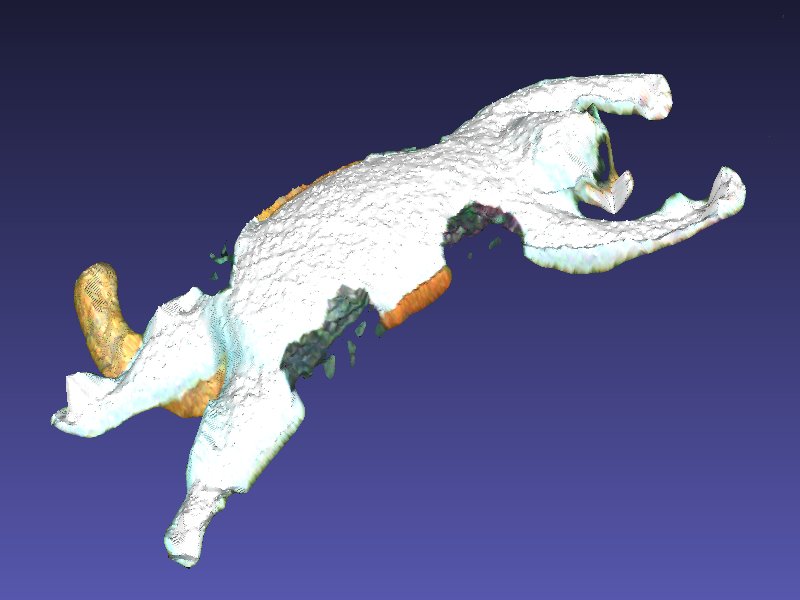}
    \caption{Reconstructed 3D shape}
    \label{fig:ststereo}
\end{figure}

Figures \ref{fig:chessboard} shows a captured image of a chessboard, and Figure \ref{fig:real_chess} shows the mirror normals and distances calibrated by the proposed method and the conventional methods. While the estimated mirror parameters look close to each other, the reprojection errors $E_\mathrm{rep}$ of the proposed, the baseline, and Takahashi \etal were 3.37, 4.75, and 13.6 pixels respectively. These reprojection errors are higher than simulation results and this is because of the localization accuracy of corresponding points and nonplanarity of mirrors. Figure \ref{fig:ststereo} shows a 3D rendering of the estimated 3D shape using the mirror parameters calibrated by the proposed method, while the residual reprojection error indicates the parameters can be further improved for example through the 3D shape reconstruction process itself\cite{furukawa2009accurate}.

From these results, we can conclude that the proposed method performs reasonably and provides a sufficiently accurate calibration for 3D shape reconstruction.

% \subsection{Discussions}

% \paragraph{Degeneracy of \cite{takahashi12new}}

% As discussed in \cite{takahashi12new}, the orthogonality constraint degenerates if a vector $\VEC{p}_i^{(l)} - \VEC{p}_j^{(l)}$ and another vector $\VEC{p}_i^{(l')} - \VEC{p}_j^{(l')}$ used in Eq \eqref{eq:orthogonality} are parallel. This indicates that in the kaleidoscopic configuration as shown in Figure \ref{fig:chessboard}, chess corner pairs between two boards cannot provide sufficiently independent constraints.

% Also while it is not explicitly discussed in \cite{takahashi12new}, the method degenerates obviously if $\VEC{m}_{12}$, $\VEC{m}_{23}$ and $\VEC{m}_{31}$ are parallel since the normal is recovered by $\VEC{n}_i = \VEC{m}_{ij} \times \VEC{m}_{ki}$.

% From these two difficulties, \cite{takahashi12new} could not perform properly for the kaleidoscopic configuration.

\section{Conclusion}

This paper proposed a new linear calibration of kaleidoscopic mirror system from 2D kaleidoscopic projections of a single 3D point in the scene. The key point to realize our linear method is to utilize not 3D positions of multiple reflections but their 2D projections.

One of the advantages of our approach is the fact that the proposed method does not require knowing the 3D geometry of the 2D points for calibration, while the conventional methods require 2D-to-3D correspondences. This indicates that our method can utilize 3D points on the target object surface of unknown geometry, and this point is verified by the evaluations in which the proposed method with non-planar calibration points outperforms the conventional methods even without bundle adjustment.

Inversely, our method assumes the 2D correspondences are given a priori. This is not a trivial problem\cite{reshetouski13discovering}, and integration with such automatic correspondence search and chamber segmentation should be further investigated to realize a complete calibration procedure for kaleidoscopic imaging system.

% Our future work includes automatic chamber segmentation\cite{reshetouski13discovering} using the proposed single point calibration, and self-calibration including the projector as well.

\section*{Acknowledgement}
This research is partially supported by JSPS Kakenhi Grant Number 26240023.

{\small
\bibliographystyle{ieee}
\bibliography{egpaper_for_review}

\begin{thebibliography}{10}\itemsep=-1pt

\bibitem{agrawal13extrinsic}
A.~Agrawal.
\newblock Extrinsic camera calibration without a direct view using spherical
  mirror.
\newblock In {\em Proc.\ of ICCV}, 2013.

\bibitem{forbes06shape}
K.~Forbes, F.~Nicolls, G.~D. Jager, and A.~Voigt.
\newblock Shape-from-silhouette with two mirrors and an uncalibrated camera.
\newblock In {\em Proc.\ of ECCV}, 2006.

\bibitem{fuchs12design}
M.~Fuchs, M.~K{\"a}chele, and S.~Rusinkiewicz.
\newblock Design and fabrication of faceted mirror arrays for light field
  capture.
\newblock In {\em Workshop on Vision, Modeling and Visualization}, 2012.

\bibitem{furukawa2009accurate}
Y.~Furukawa and J.~Ponce.
\newblock Accurate camera calibration from multi-view stereo and bundle
  adjustment.
\newblock {\em IJCV}, 84(3):257--268, 2009.

\bibitem{gluckman2001catadioptric}
J.~Gluckman and S.~K. Nayar.
\newblock Catadioptric stereo using planar mirrors.
\newblock {\em IJCV}, 44(1):65--79, 2001.

\bibitem{goshtasby93design}
A.~Goshtasby and W.~A. Gruver.
\newblock Design of a single-lens stereo camera system.
\newblock {\em Pattern Recognition}, 26(6):923 -- 937, 1993.

\bibitem{hartley00multiple}
R.~I. Hartley and A.~Zisserman.
\newblock {\em Multiple View Geometry in Computer Vision}.
\newblock Cambridge University Press, 2000.

\bibitem{hesch08mirror}
J.~A. Hesch, A.~I. Mourikis, and S.~I. Roumeliotis.
\newblock {\em Algorithmic Foundation of Robotics {VIII}}, volume~57 of {\em
  Springer Tracts in Advanced Robotics}, chapter Mirror-Based Extrinsic Camera
  Calibration, pages 285--299.
\newblock 2009.

\bibitem{huang06contour}
P.-H. Huang and S.-H. Lai.
\newblock Contour-based structure from reflection.
\newblock In {\em Proc.\ of CVPR}, volume~1, pages 379--386, 2006.

\bibitem{ihrke2012kaleidoscopic}
I.~Ihrke, I.~Reshetouski, A.~Manakov, A.~Tevs, M.~Wand, and H.-P. Seidel.
\newblock A kaleidoscopic approach to surround geometry and reflectance
  acquisition.
\newblock In {\em CVPR Workshop on Computational Cameras and Displays}, pages
  29--36, 2012.

\bibitem{inoshita13full}
C.~Inoshita, S.~Tagawa, M.~A. Mannan, Y.~Mukaigawa, and Y.~Yagi.
\newblock Full-dimensional sampling and analysis of bssrdf.
\newblock {\em IPSJ Transactions on Computer Vision and Applications},
  5:119--123, 2013.

\bibitem{kumar08simple}
R.~Kumar, A.~Ilie, J.-M. Frahm, and M.~Pollefeys.
\newblock Simple calibration of non-overlapping cameras with a mirror.
\newblock In {\em Proc.\ of CVPR}, pages 1--7, 2008.

\bibitem{lanman09surround}
D.~Lanman, D.~Crispell, and G.~Taubin.
\newblock Surround structured lightning: 3-d scanning with orthographic
  illumination.
\newblock In {\em CVIU}, pages 1107--1117, November 2009.

\bibitem{lepetit08epnp}
V.~Lepetit, F.~Moreno-Noguer, and P.~Fua.
\newblock Epnp: An accurate o(n) solution to the pnp problem.
\newblock {\em IJCV}, 81(2), 2008.

\bibitem{levoy04synthetic}
M.~Levoy, B.~Chen, V.~Vaish, M.~Horowitz, I.~McDowall, and M.~Bolas.
\newblock Synthetic aperture confocal imaging.
\newblock In {\em Proc.\ of SIGGRAPH}, pages 825--834, 2004.

\bibitem{long15simplified}
G.~Long, L.~Kneip, X.~Li, X.~Zhang, and Q.~Yu.
\newblock Simplified mirror-based camera pose computation via rotation
  averaging.
\newblock In {\em Proc.\ of CVPR}, pages 1247--1255, 2015.

\bibitem{nane98stereo}
S.~A. Nene and S.~K. Nayar.
\newblock Stereo with mirrors.
\newblock In {\em Proc.\ of ICCV}, pages 1087--1094, 1998.

\bibitem{nobuhara16single}
S.~Nobuhara, T.~Kashino, T.~Matsuyama, K.~Takeuchi, , and K.~Fujii.
\newblock A single-shot multi-path interference resolution for mirror-based
  full 3d shape measurement with a correlation-based tof camera.
\newblock In {\em Proc.\ of 3DV}, 2016.

\bibitem{reshetouski13discovering}
I.~Reshetouski, A.~M. iand Ayush~Bhandari, R.~Raskar, H.-P. Seidel, and
  I.~Ihrke.
\newblock Discovering the structure of a planar mirror system from multiple
  observations of a single point.
\newblock In {\em Proc.\ of CVPR}, pages 89--96, 2013.

\bibitem{reshetouski2013mirrors}
I.~Reshetouski and I.~Ihrke.
\newblock {\em Mirrors in Computer Graphics, Computer Vision and Time-of-Flight
  Imaging}, pages 77--104.
\newblock Springer Berlin Heidelberg, 2013.

\bibitem{reshetouski11three}
I.~Reshetouski, A.~Manakov, H.-P. Seidel, and I.~Ihrke.
\newblock Three-dimensional kaleidoscopic imaging.
\newblock In {\em Proc.\ of CVPR}, pages 353--360, 2011.

\bibitem{rodorigues10camera}
R.~Rodrigues, P.~Barreto, and U.~Nunes.
\newblock Camera pose estimation using images of planar mirror reflections.
\newblock In {\em Proc.\ of ECCV}, pages 382--395, 2010.

\bibitem{sen05dual}
P.~Sen, B.~Chen, G.~Garg, S.~R. Marschner, M.~Horowitz, M.~Levoy, and H.~P.~A.
  Lensch.
\newblock Dual photography.
\newblock In {\em Proc.\ of SIGGRAPH}, pages 745--755, 2005.

\bibitem{sturm06how}
P.~Sturm and T.~Bonfort.
\newblock How to compute the pose of an object without a direct view.
\newblock In {\em Proc.\ of ACCV}, pages 21--31, 2006.

\bibitem{tagawa12eight}
S.~Tagawa, Y.~Mukaigawa, and Y.~Yagi.
\newblock 8-d reflectance field for computational photography.
\newblock In {\em Proc.\ of ICPR}, pages 2181--2185, 2012.

\bibitem{tahara15interference}
T.~Tahara, R.~Kawahara, S.~Nobuhara, and T.~Matsuyama.
\newblock Interference-free epipole-centered structured light pattern for
  mirror-based multi-view active stereo.
\newblock In {\em Proc.\ of 3DV}, pages 153--161, 2015.

\bibitem{takahashi12new}
K.~Takahashi, S.~Nobuhara, and T.~Matsuyama.
\newblock A new mirror-based extrinsic camera calibration using an
  orthogonality constraint.
\newblock In {\em Proc.\ of CVPR}, pages 1051--1058, 2012.

\bibitem{ying13self}
X.~Ying, K.~Peng, Y.~Hou, S.~Guan, J.~Kong, and H.~Zha.
\newblock Self-calibration of catadioptric camera with two planar mirrors from
  silhouettes.
\newblock {\em TPAMI}, 35(5):1206--1220, 2013.

\bibitem{zhang1998}
Z.~Zhang.
\newblock A flexible new technique for camera calibration.
\newblock {\em TPAMI}, 22:1330--1334, 1998.

\end{thebibliography}
}

\end{document}